\documentclass[runningheads]{llncs}

\usepackage{accv}

\usepackage{accvabbrv}
\usepackage{graphicx}
\usepackage{booktabs}
\usepackage{amsmath}
\usepackage{amssymb}
\usepackage{multirow}
\usepackage{xcolor}
\usepackage{colortbl}
\usepackage{xspace}
\usepackage[accsupp]{axessibility}
\usepackage{placeins}
 
\usepackage{makecell}

\usepackage{hyperref}

\usepackage{orcidlink}

\usepackage[capitalize]{cleveref}
\usepackage[all]{hypcap}
\Crefname{figure}{Fig.}{Figs.}
\Crefname{figure}{Fig.}{Figs.}
\Crefname{section}{Sec.}{Secs.}
\Crefname{section}{Sec.}{Secs.}
\Crefname{subsection}{Sec.}{Secs.}
\Crefname{subsection}{Sec.}{Secs.}
\Crefname{equation}{Eq.}{Eqs.}
\Crefname{equation}{Eq.}{Eqs.}
\Crefname{table}{Table}{Tables}
\Crefname{table}{Table}{Tables}
\Crefname{appendix}{Appendix}{Appendices}
\Crefname{appendix}{Appendix}{Appendices}

\newcommand{\methodname}{VGGT-Bridge\xspace}
 \newcommand{\SimT}{\mathrm{Sim}(3)}     

\begin{document}

\title{VGGT-Bridge: Beyond Sequential Pose Graphs via Coarse-Stride Skip Edges}

\titlerunning{VGGT-Bridge: Coarse-Stride Skip Edges}

\author{Sungjae Choi\thanks{Equal contribution.} \and
Hanna Bae$^{\star}$ \and
Sunghyun Baek \and
Junmo Kim}
\authorrunning{S. Choi et al.}
\institute{Korea Advanced Institute of Science and Technology, South Korea\\
\email{\{sungjae579, hannabae, baeksh, junmo.kim\}@kaist.ac.kr}}

\maketitle
\begin{abstract}

Feed-forward visual geometry transformers such as VGGT reconstruct dense 3D structure from images in a single forward pass, simplifying multi-view 3D reconstruction.
However, their quadratic attention complexity makes them difficult to scale to long sequences with thousands of frames.
Chunk-and-align frameworks address this by splitting a long sequence into overlapping chunks and stitching their local reconstructions into a pose graph.
Yet existing methods connect only sequentially adjacent chunks, so small per-frame errors accumulate along the chain into large-scale drift.
To move beyond sequential edges, we propose \methodname, which adds long-range skip edges that directly constrain non-adjacent chunks without retraining.
By running VGGT on sparsely sampled coarse chunks, each coarse chunk bridges distant fine chunks into a single direct constraint.
We further turn VGGT's first-frame scale bias into a drift correction by feeding selected coarse chunks in reverse, and a loop-aware policy keeps this reversal compatible with existing loop closures.
\methodname reduces ATE by 28.3\% on KITTI Odometry, 18.8\% on Virtual KITTI, and 10.0\% on Waymo Open over the SwiftVGGT baseline, achieving the best performance among all chunk-and-align methods.
Our code is available at \url{https://github.com/BeCow5X5/VGGT-Bridge}.

\end{abstract}

\section{Introduction}
\label{sec:intro}

\begin{figure}[h]
\centering
\includegraphics[width=\linewidth]{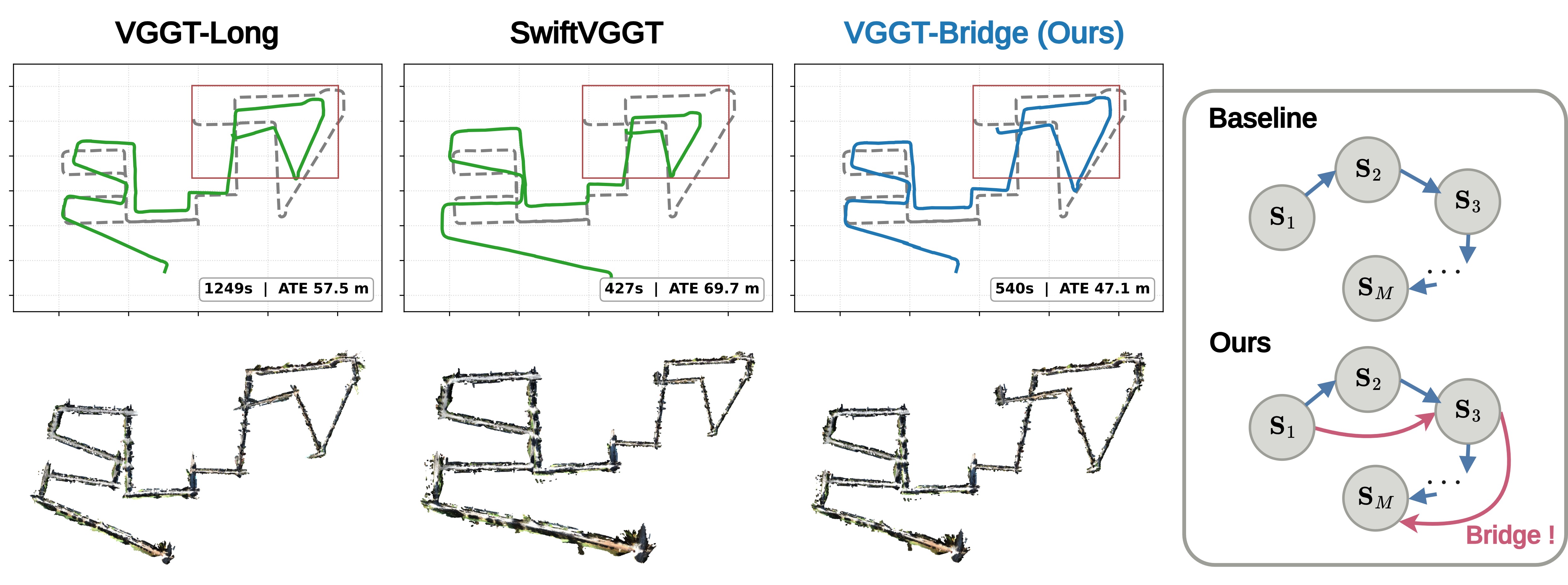}\\
\caption{Prior methods connect only sequentially adjacent chunks ($\mathbf{S}_1\!\to\!\mathbf{S}_2\!\to\!\cdots$), so per-frame errors accumulate along the chain into large-scale drift.
\methodname inserts \emph{skip edges} (pink arcs, \eg, $\mathbf{S}_1\!\to\!\mathbf{S}_3$) that directly constrain non-adjacent chunks under the chunk-and-align pipeline.
On KITTI Odometry Seq.~08, these long-range constraints lower ATE RMSE to $47.1$\,m.}
\label{fig:intro}
\end{figure}

Feed-forward visual geometry transformers such as VGGT~\cite{vggt}, DUSt3R~\cite{dust3r}, and MASt3R~\cite{mast3r} predict dense per-frame depth, camera pose and 3D structure directly from a sequence of images in a single forward pass.
They have dramatically simplified multi-view 3D reconstruction by eliminating feature matching, bundle adjustment, and iterative refinement.
Many real-world applications, such as autonomous driving and large-scale mapping, require reconstructing long video sequences with thousands of frames, yet these models still struggle to scale because their attention complexity grows quadratically with the number of input frames.
To scale these models to long sequences, a common chunk-and-align pipeline~\cite{vggtlong,swiftvggt} splits a long sequence into overlapping chunks, runs the feed-forward model on each chunk independently, and aligns neighboring chunks through the frames shared in their overlaps.

However, these methods connect only temporally adjacent chunks, so two chunks more than one step apart are related solely through the chain of intermediate alignments.
Small errors in these local alignments propagate along the chain, and even after global pose graph optimization they accumulate into large-scale drift.
Also, VGGT anchors each chunk to its first frame as a reference and reconstructs later frames at a progressively smaller scale.
This anchoring bias worsens the drift, since the shrinkage acts in one direction and its errors accumulate rather than cancel.
When a reliable revisit is detected, loop closure can supply occasional non-local constraints, yet such edges are sparse and may be absent in non-revisiting sequences, such as long, loop-free highway driving.

Motivated by this, we propose \methodname, which adds training-free long-range constraints to the same lightweight pose graph, without retraining the model or using external modules.
\methodname consists of three components.
First, we augment the sequential chain with long-range \emph{skip edges} between non-adjacent chunks.
Running the same model on coarse chunks, which sample frames more sparsely and span a wider temporal range, each coarse chunk bridges and directly constrains the distant fine chunks it overlaps.
Second, to correct the anchoring bias, we reverse the input frame order of selected coarse chunks, which shifts their scale against the bias and turns the skip edge into a corrective anti-bias signal for the constructed pose graph.
Finally, this reversal can conflict with the loop closures that already correct scale drift in loop-rich regions.
We therefore introduce a policy that reverses only where drift dominates and stays forward where reliable loops are present.

We extensively validate \methodname on three driving benchmarks, KITTI Odometry~\cite{kitti_odometry}, Waymo Open Dataset~\cite{waymo}, and Virtual KITTI~\cite{vkitti}, where it attains the best performance among all chunk-and-align methods.
On KITTI Odometry, \methodname lowers the Absolute Trajectory Error (ATE) from $31.53$\,m to $22.61$\,m over the SwiftVGGT baseline, with the largest gains on drift-dominated sequences.
Beyond the overall accuracy, we directly verify the underlying mechanism, showing that reverse anchoring shifts the scale of skip edges in the direction opposite to the bias and that this anti-bias signal suppresses global scale drift.
Furthermore, we provide component ablations and an efficiency analysis that support the contribution of each design choice.
In summary, our key contributions are as follows:
\begin{itemize}
\item[$\bullet$] We propose \methodname, which links non-adjacent chunks with skip edges to add long-range constraints to the chunk-and-align pipeline, without any training or external module.

\item[$\bullet$] We correct the first-frame anchoring bias of feed-forward models by reversing the input order of selected coarse chunks, with a loop-aware policy setting each chunk's direction to stay compatible with real loop closures.

\item[$\bullet$] Through extensive experiments on KITTI Odometry, Waymo Open Dataset, and Virtual KITTI, we show that \methodname achieves the best performance among all chunk-and-align methods, and validate its design through analyses of the scale bias and component ablations.

\end{itemize}

\section{Related Work}
\label{sec:related}

\textbf{Feed-forward visual geometry models.}
Classical 3D reconstruction relies on modular pipelines for correspondence estimation, camera recovery, triangulation, and bundle adjustment~\cite{mvg_book,rome,colmap,mvs,robustsfm,globalsfm}.
Learning-based SfM methods improve or differentiate parts of this pipeline~\cite{banet,deepsfm,droidslam,flowmap,vggsfm}, but still retain explicit geometric estimation and optimization.
Feed-forward visual geometry models take a more direct route, predicting dense 3D structure from images with pretrained networks.
Beginning with pairwise point map prediction~\cite{dust3r,mast3r}, recent systems extend this paradigm with spatial memory, unconstrained SfM, or multi-view reconstruction~\cite{spann3r,mast3rsfm,fast3r,pi3,vggt}.
While these models provide strong short-window geometry, their global attention scales quadratically with the number of frames, making long sequences prohibitively expensive.

\textbf{Classical long-scene reconstruction.}
Long-scene reconstruction has traditionally been addressed by simultaneous localization and mapping (SLAM) and incremental structure-from-motion (SfM), where a visual front-end establishes short-range geometric constraints from image measurements and a geometric back-end refines camera poses and map structure over time~\cite{ptam,lsdslam,orbslam2015,orbslam2,orbslam3,dso,badslam,droidslam,dpvo}.
These systems combine local tracking or odometry with map-level optimization, and reduce long-term drift by adding non-local constraints from loop detection and visual place recognition~\cite{dinov2,salad}.
The resulting loop edges are incorporated through pose graph or factor graph optimization backends~\cite{graphslam,g2o,gtsam}.
This pipeline has been highly effective for long sequences, but it relies on a carefully designed stack of correspondence estimation, mapping, place recognition, and optimization.
Recent feed-forward approaches pursue the same goal of long-scene consistency while replacing much of this hand-designed front-end with learned dense geometry priors.

\textbf{Extending feed-forward models to long scenes.}
One line of work modifies the model or inference process itself, using spatial memory, recurrent states, streaming formulations, or test-time training to extend the temporal context available to the network~\cite{point3r,long3r,cut3r,ttt3r,stream3r,streamvggt,longstream,loger,scal3r}.
These approaches broaden the usable context, but often require retraining, architectural changes, or a streaming reformulation.
A complementary direction keeps the pretrained model fixed and moves long-scene consistency outside the network by constructing a graph over local reconstructions~\cite{mast3rslam,vggtslam,vggtslam2,talo}.
Chunk-and-align pipelines follow this route. 
They split a long sequence into overlapping chunks, run the feed-forward model on each, and fuse the local reconstructions in a global pose graph.
VGGT-Long~\cite{vggtlong} aligns neighboring chunks with Iteratively Reweighted Least Squares (IRLS) over Sim(3) and adds loop closures from visual place recognition, while SwiftVGGT~\cite{swiftvggt} replaces the iterative alignment with a single-shot Umeyama SVD~\cite{umeyama} step and reuses the model's internal descriptors for loop detection.
VGGT-Motion~\cite{vggtmotion} uses motion-aware submaps to reduce redundant VGGT processing and improve submap alignment.
Despite these differences, existing graph-based approaches remain largely sequential, with most constraints coming from adjacent chunk overlaps and non-local edges mainly arising from loop closures.
Our work instead introduces additional long-range constraints that connect distant chunks beyond the sequential overlap chain. 

\section{Method}
\label{sec:method}
\subsection{Overview}
\label{sec:method_overview}
We first review the chunk-and-align pipeline that our method builds on, which reconstructs a long sequence by splitting it into chunks and stitching their local reconstructions into a single pose graph.
We then give an overview of \methodname, which augments this pose graph with training-free long-range edges to reduce the drift that accumulates along the sequence.
\begin{figure}[t]
\centering
\includegraphics[width=\linewidth]{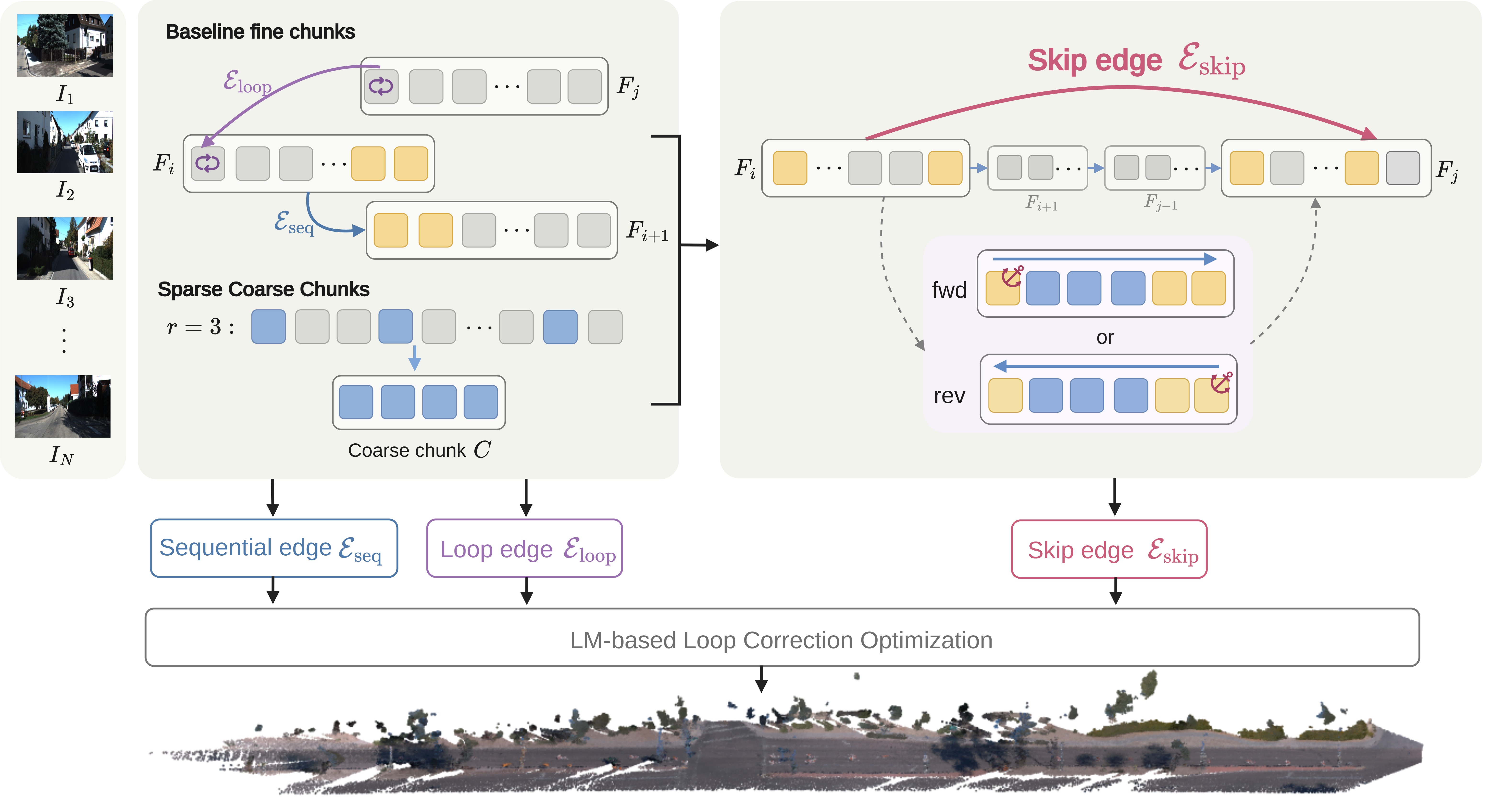}
\caption{\textbf{Method overview.}
Gray, blue, and yellow boxes denote input frames, sparsely sampled coarse frames, and alignment overlaps, respectively.
Standard chunk-and-align pipelines construct overlapping fine chunks and form sequential edges $\mathcal{E}_{\mathrm{seq}}$ (blue) between adjacent chunks, with optional loop edges $\mathcal{E}_{\mathrm{loop}}$ (purple) between revisited chunks.
VGGT-Bridge additionally runs VGGT on sparse coarse chunks and composes two cross-scale $\mathrm{Sim}(3)$ alignments through each coarse chunk into a long-range skip edge $\mathcal{E}_{\mathrm{skip}}$ (rose) between non-adjacent fine chunks.
Forward/reverse anchor markers show the selected processing direction of each coarse chunk, chosen by a loop-aware policy to complement the available loop closures.
All edge sets are jointly optimized in a lightweight $\mathrm{Sim}(3)$ pose graph to recover globally consistent chunk poses.}
\label{fig:method}
\end{figure}

\subsubsection{Chunk-and-Align Pipeline}
Given an image sequence $\{I_1,\dots,I_N\}$, a chunk-and-align pipeline splits it into overlapping fine chunks $F_1,\dots,F_M$, each spanning $B$ contiguous frames with $O$ overlapping frames.
It runs VGGT~\cite{vggt} on each chunk independently and obtains per-frame extrinsics, depth, and 3D points expressed in the chunk's local coordinate system.
For each chunk, the first input image is assigned identity extrinsics and serves as the reference frame.
A pairwise $\mathrm{Sim}(3)$ alignment is then computed between each pair of consecutive chunks $F_i, F_{i+1}$ from their $O$ overlapping frames, yielding the relative transformation $\mathbf{S}_{i,i+1}\in\mathrm{Sim}(3)$, parametrized by a scale $s\in\mathbb{R}_{>0}$, a rotation $\mathbf{R}\in\mathrm{SO}(3)$, and a translation $\mathbf{t}\in\mathbb{R}^3$.
To correct the drift these sequential alignments accumulate, it further links revisited chunks $F_i, F_j$ with loop-closure alignments $\mathbf{S}_{ij}^{\mathrm{loop}}$, where revisits are found by place recognition and aligned through a shared loop-centric chunk.
%
Finally, pose graph optimization recovers the chunk poses in a common frame, treating each pose as a node and each measured transform as an edge that constrains the two poses it connects.
It minimizes the mismatch between every measurement and the relative pose it implies,
\begin{equation}
\{\mathbf{S}_i^\star\}
\;=\;
\arg\min_{\{\mathbf{S}_i\}}\;
\sum_{i}
\bigl\|\log\!\bigl(\mathbf{S}_{i, i+1}^{-1}\mathbf{S}_i^{-1}\mathbf{S}_{i+1}\bigr)\bigr\|^2
\;+\!\!
\sum_{(i,j)}
\bigl\|\log\!\bigl((\mathbf{S}_{ij}^{\mathrm{loop}})^{-1}\mathbf{S}_i^{-1}\mathbf{S}_j\bigr)\bigr\|^2,
\label{eq:pgo}
\end{equation}
solved with Levenberg--Marquardt~\cite{levenberg,marquardt} on the $\mathrm{Sim}(3)$ manifold.

\subsubsection{VGGT-Bridge}
We augment the pose graph with long-range edges through three components, leaving both the feed-forward model and the optimizer unchanged, as illustrated in \Cref{fig:method}.
First, we run VGGT on coarse-stride chunks that each overlap several fine chunks, and compose their alignments into a \emph{skip edge} directly constraining two non-adjacent fine chunks, as detailed in \Cref{sec:method_skipedge}.
Second, we identify a first-frame anchor bias in VGGT and feed selected coarse chunks in reverse temporal order, turning the skip edge into a \emph{pseudo loop closure} whose reverse-coarse constraint counteracts the accumulated scale drift, developed in \Cref{sec:method_anchor}.
Third, a loop-aware policy selects the coarse anchor position per chunk, reconciling the skip edges with loop closures in loop-rich regions, which we describe in \Cref{sec:method_adaptive}.
All edges, sequential, loop, and skip, are then optimized jointly in a single Sim(3) pose graph, as formalized in \Cref{sec:method_fin}.


\subsection{Coarse-stride skip edges}
\label{sec:method_skipedge}
We denote each edge by the measured relative transform between two chunk nodes, written $\mathbf{S}_{ij}\in\SimT$ for the edge from chunk $i$ to chunk $j$.
The sequential alignments form the edge set
\begin{equation}
\mathcal{E}_{\mathrm{   seq}} = \bigl\{\mathbf{S}_{i, i+1} \;:\; i=1,\dots,M-1\bigr\}.
\end{equation}
Since $\mathcal{E}_{\mathrm{seq}}$ has no direct edge between non-adjacent chunks, their relative pose is only recovered implicitly from the composed intermediate edges, letting the error of each edge accumulate along the path.
To add long-range constraints, we apply the same model to coarse-stride chunks, which sample frames more sparsely and cover a wider temporal extent.
For a stride $r\in\mathcal{R}$, we subsample the sequence to every $r$-th frame and partition it into coarse chunks $\mathcal{C}^{r}=\{C_1^{r},\dots,C_{M_r}^{r}\}$ of $B^{r}$ frames each.
Each coarse chunk $C_k^{r}$ then spans $r\cdot B^{r}$ original frames, wider than the $B$ frames of a fine chunk.
For instance, while a fine chunk spans $75$ frames, a stride-3 coarse chunk with $B^{3}{=}35$ frames spans $\approx105$ original frames, covering a wider temporal range.
As with fine chunks, VGGT produces per-frame extrinsics, depth, and 3D points in each coarse chunk's local frame.

A coarse chunk $C_k^{r}$ and a fine chunk $F_i$ share frames when those frames occur as subsampled positions in $C_k^{r}$ and as contiguous positions in $F_i$.
From the 3D points on these shared frames, we keep a reliable subset whose depth discrepancy between the two chunks is small and whose predicted confidence is high.
A single Umeyama SVD~\cite{umeyama} step on this subset then estimates the fine-to-coarse alignment $S_{F_i, C_k^{r}}\in\SimT$ between $F_i$ and $C_k^{r}$.
The sequential edges of $\mathcal{E}_{\mathrm{seq}}$ are estimated the same way, from the 3D points on the overlap between adjacent fine chunks.
A single coarse chunk overlaps several consecutive fine chunks, but linking every one of them to it would over-constrain the graph.
We therefore connect only the farthest pair, the lowest and highest indexed fine chunks $F_{\mathrm{min}}$ and $F_{\mathrm{max}}$ overlapping $C_k^{r}$.
When the two are non-adjacent, that is, $\mathrm{idx}(F_{\mathrm{max}})-\mathrm{idx}(F_{\mathrm{min}})\geq 2$, we chain their two fine-to-coarse alignments $S_{F_i, C_k^{r}}$ and $S_{F_j, C_k^{r}}$ through the shared coarse chunk $C_k^{r}$ into a single skip edge,
\begin{equation}
\mathbf{S}_{ij}^{\mathrm{skip}}
\;=\;
\mathbf{S}_{F_i, C_k^{r}}\;\circ\;\mathbf{S}_{F_j, C_k^{r}}^{-1}
\;\in\;\SimT,\quad
(i,j)=(\mathrm{idx}(F_{\mathrm{min}}),\mathrm{idx}(F_{\mathrm{max}})),
\label{eq:skip}
\end{equation}
which is a direct $\SimT$ constraint from fine chunk $i$ to fine chunk $j$, obtained through the shared coarse chunk $C_k^{r}$.
Forming one such edge per coarse chunk over all strides gives the skip edge set
\begin{equation}
\mathcal{E}_{\mathrm{skip}}
\;=\;
\bigl\{\mathbf{S}_{ij}^{\mathrm{skip}} \;:\; k=1,\dots,M_r,\; r\in\mathcal{R}\bigr\},
\end{equation}
where $\mathcal{R}\subset\mathbb{N}_{>1}$ is the set of coarse strides we use.
This skip edge gives non-adjacent fine chunks a wider, explicit constraint, so that they no longer rely on the implicit sequential chain.


\subsection{Reverse-anchored coarse chunks}
\label{sec:method_anchor}
VGGT anchors each chunk at its first input frame, and this anchoring biases the reconstructed scale to shrink away from that anchor as the input advances.
Since every chunk is anchored the same way, this bias always acts in the same direction, so the scale drift accumulates without canceling along the sequential edges.
A loop closure would correct this drift by supplying the pose graph with a constraint from a revisited place, but such a constraint is available only where a revisit is detected, leaving the drift uncorrected in loop-free regions.
In these regions, we repurpose the skip edge as a pseudo loop closure, a non-local constraint that corrects drift in place of a real loop closure.

A default skip edge is built from coarse chunks that are anchored at their first frame, just like the fine chunks, so it carries the same first-frame bias and cannot counteract the drift it shares with the sequential edges.
We reverse this bias by feeding the frames of the coarse chunk that forms the skip edge in reverse temporal order.
The forward-anchored chunk $C_k^{r,\mathrm{fwd}}$, used everywhere so far, feeds its frames $(f_1,\dots,f_{B^r})$ in temporal order and is anchored at the earliest frame $f_1$.
The reverse-anchored chunk $C_k^{r,\mathrm{rev}}$ feeds the same frames in reverse order $(f_{B^r},\dots,f_1)$, moving the anchor to the latest frame $f_{B^r}$, and its outputs are then reordered to temporal order for alignment.

We call the skip edge composed from a forward-anchored chunk a forward skip edge, and the one composed from a reverse-anchored chunk a reverse skip edge.
Reversing the input leaves the composition of \Cref{eq:skip} unchanged but moves the anchor from the first frame to the last, so the scale bias now acts in the opposite direction.
The reverse skip edge therefore introduces a scale constraint in the direction opposite to the accumulated forward drift, acting as an anti-bias constraint.

\subsection{Loop-aware adaptive direction policy}
\label{sec:method_adaptive}
The reversal above targets loop-free regions, but a sequence also contains loop-rich ones, where the loop edges in loop edge set $\mathcal{E}_{\mathrm{loop}}$ already constrain the drift.
There, a reverse skip edge would pull against these loop edges, so we keep loop-rich regions forward and reverse only where loops are absent.
Assigning a single direction to all coarse chunks of each stride cannot adapt to both region types, so we choose the direction per coarse chunk from the local loop density.
A detected loop links two revisiting frames, found by place recognition~\cite{swiftvggt}, and we write $\mathcal{P}$ for the set of these loop frame pairs $(a,b)$.
From these pairs we define a per-frame loop density
\begin{equation}
\rho(t)
\;=\;
\frac{1}{\rho_{\max}}\sum_{(a,b)\in\mathcal{P}}\bigl[
\mathbf{1}\{|t-a|<W\}+\mathbf{1}\{|t-b|<W\}
\bigr],
\label{eq:loop_density}
\end{equation}
which counts how many loop frames fall within a window of $W$ frames around frame $t$.
Here $\rho_{\max}$ is the maximum of this unnormalized count over all frames, so that $\rho\in[0,1]$.
We apply the adaptive choice to a subset of strides $\mathcal{R}_a\subseteq\mathcal{R}$ and keep the remaining strides forward.
For each coarse chunk $C_k^{r}$ of a stride $r\in\mathcal{R}_a$, we average the density over its frames,
\begin{equation}
\bar\rho_k^{r}=\frac{1}{|C_k^{r}|}\sum_{t\in C_k^{r}}\rho(t),
\end{equation}
and use it to set the chunk's anchor position.
If \(\bar\rho_k^{r}\) falls below the threshold $\tau$, the chunk lies in a loop-sparse region and we make it a reverse-anchored chunk.
Otherwise it lies in a loop-rich region and we keep it a forward-anchored chunk.
Each adaptive coarse chunk is thus anchored according to its local regime, so that reverse-anchored chunks supply the anti-bias signal and forward-anchored chunks do not compete with nearby loop edges.

\subsection{Global optimization}
\label{sec:method_fin}
The three edge sets are combined into a single $\SimT$ pose graph over $\mathcal{E}=\mathcal{E}_{\mathrm{seq}}\cup\mathcal{E}_{\mathrm{loop}}\cup\mathcal{E}_{\mathrm{skip}}$ and solved jointly,
\begin{equation}
\{\mathbf{S}_i^\star\}
\;=\;
\arg\min_{\{\mathbf{S}_i\}}\;
\sum_{\mathbf{S}_{ij}\in\mathcal{E}}
w_{ij}\,
\bigl\|\log\!\bigl(\mathbf{S}_{ij}^{-1}\mathbf{S}_i^{-1}\mathbf{S}_j\bigr)\bigr\|^2,
\qquad
w_{ij}=\bar w_{ij}\,\gamma_{ij},
\label{eq:pgo_full}
\end{equation}
where $\log(\cdot)$ maps a relative $\SimT$ transform to its $7$-D Lie algebra representation.
Each edge weight $w_{ij}$ factors into a fixed base weight $\bar w_{ij}$, set per edge type when the edge is formed, and a residual-dependent factor $\gamma_{ij}$.
The residual $\mathbf{r}_{ij}=\log\!\bigl(\mathbf{S}_{ij}^{-1}\mathbf{S}_i^{-1}\mathbf{S}_j\bigr)$ is the same $\SimT$ discrepancy minimized in \Cref{eq:pgo_full}.
To make the joint solution robust to occasional unreliable skip edges, we recompute $\gamma_{ij}$ from this residual at every iteration, so that a large residual lowers $\gamma_{ij}$ toward zero.
We implement $\gamma_{ij}$ as a Cauchy factor $1/(1+\|\mathbf{r}_{ij}\|^2/\delta^2)$ with $\delta{=}2.0$, which keeps a few outlier skip or loop edges from distorting the global solution.
We solve \Cref{eq:pgo_full} with Levenberg--Marquardt, leaving the optimizer unchanged.

\section{Experiments}
\label{sec:exp}

\subsection{Setup}
\label{sec:exp_setup}
\noindent\textbf{Datasets.} 
We evaluate our method on three autonomous driving benchmarks: KITTI Odometry~\cite{kitti_odometry}, Waymo Open Dataset (v1.4.1)~\cite{waymo}, and Virtual KITTI  (v1.3.1)~\cite{vkitti}. Specifically, for the Virtual KITTI dataset, we evaluate five synthetic driving scenes under all six photometric variants.

\noindent\textbf{Baselines.} 
We compare against two groups of methods. The first consists of \emph{calibrated} SLAM baselines that require ground-truth camera intrinsics: ORB-SLAM2~\cite{orbslam2}, DPVO~\cite{dpvo}, and DROID-SLAM~\cite{droidslam}, which serve as an upper-bound reference. The second consists of \emph{calibration-free} methods that operate without intrinsics: MASt3R-SLAM~\cite{mast3rslam}, CUT3R~\cite{cut3r}, Fast3R~\cite{fast3r}, VGGT-SLAM~\cite{vggtslam}, VGGT-Motion~\cite{vggtmotion}, VGGT-Long~\cite{vggtlong}, and SwiftVGGT~\cite{swiftvggt}. Among these, SwiftVGGT, VGGT-Long, and our method share the same VGGT checkpoint and chunk construction, and we reproduce them for a fair comparison.\footnote{The remaining numbers are taken from~\cite{vggtmotion}.}

\noindent\textbf{Metric.}
We report Sim(3) Umeyama-aligned translation Absolute Trajectory Error (ATE) RMSE (m), following VGGT-Long~\cite{vggtlong}. 
The full-transformation metric used in the SwiftVGGT~\cite{swiftvggt} paper is reported in Supp.

\noindent\textbf{Implementation Details.} 
We utilize the official VGGT checkpoint~\cite{vggt} and PCA-whitened DINO tokens for loop detection~\cite{swiftvggt}.
Following SwiftVGGT~\cite{swiftvggt}, we use a chunk size $B{=}75$ and overlap size $O{=}30$ on KITTI Odometry and Virtual~KITTI, and $B{=}60$, $O{=}24$ on Waymo.
For the skip edge sets, we set $\mathcal{R}{=}\{3, 5\}$ and apply the adaptive direction policy only to stride 3 ($\mathcal{R}_a{=}\{3\}$), keeping stride 5 always forward.
All experiments are conducted on a single NVIDIA RTX 4090, and the remaining hyperparameters are provided in Supp.

\subsection{Main Results}
\label{sec:exp_main}
We quantitatively evaluate our method on KITTI Odometry, Waymo Open, and Virtual KITTI. Across all three benchmarks, \methodname achieves the best average performance among calibration-free methods.

\noindent\textbf{KITTI Odometry.}
\begin{table}[t]
\centering
\caption{\textbf{Performance comparison on KITTI Odometry dataset.}
\textbf{Bold}/\underline{underline} mark the best/second-best among the calibration-free methods.
``Calib.'' = uses camera intrinsics, ``TL'' = tracking lost, ``OOM'' = out of memory.
Results with * are reproduced by us.
}
\label{tab:kitti_perseq}
\setlength{\tabcolsep}{2.5pt}
\renewcommand{\arraystretch}{1.1}
\scriptsize
\resizebox{\textwidth}{!}{%
\begin{tabular}{@{}l|c|rrrrrrrrrrr|r@{}}
\toprule
Method & Calib. & 00 & 01 & 02 & 03 & 04 & 05 & 06 & 07 & 08 & 09 & 10 & Avg. (\({\downarrow}\))\\
\midrule
\textit{\# frames} & & \textit{4541} & \textit{1101} & \textit{4661} & \textit{801} & \textit{271} & \textit{2761} & \textit{1101} & \textit{1101} & \textit{4071} & \textit{1591} & \textit{1201} & \\
\textit{Length (m)} & & \textit{2453} & \textit{5067} & \textit{3724} & \textit{561} & \textit{394} & \textit{2206} & \textit{1233} & \textit{650} & \textit{3223} & \textit{1705} & \textit{920} & \\
\textit{Speed (m/fr)} & & \textit{0.82} & \textit{2.23} & \textit{1.09} & \textit{0.70} & \textit{1.45} & \textit{0.80} & \textit{1.12} & \textit{0.59} & \textit{0.79} & \textit{1.07} & \textit{0.77} & \\
\textit{Loop} & & \checkmark & $\times$ & \checkmark & $\times$ & $\times$ & \checkmark & $\checkmark$ & $\checkmark$ & $\times$ & $\checkmark$ & $\times$ & \\
\midrule
ORB-SLAM2~\cite{orbslam2}        & \checkmark & 6.03 & 508.34 & 14.76 & 1.02 & 1.57 & 4.04 & 11.16 & 2.19 & 38.85 & 8.39 & 6.63 & 54.82\\
DPVO~\cite{dpvo}                 & \checkmark & 113.21 & 12.69 & 123.40 & 2.09 & 0.68 & 58.96 & 54.78 & 19.26 & 115.90 & 75.10 & 13.63 & 53.61\\
DROID-SLAM~\cite{droidslam}      & \checkmark & 92.10 & 344.60 & 107.61 & 2.38 & 1.00 & 118.50 & 62.47 & 21.78 & 161.60 & 72.32 & 118.70 & 100.28\\
\midrule
MASt3R-SLAM~\cite{mast3rslam}    & $\times$ & \multicolumn{11}{c|}{TL}  & --\\
CUT3R~\cite{cut3r}               & $\times$ & \multicolumn{11}{c|}{OOM} & --\\
Fast3R~\cite{fast3r}             & $\times$ & \multicolumn{11}{c|}{OOM} & --\\
VGGT-SLAM~\cite{vggtslam}        & $\times$ & 82.73 & 96.73 & 211.26 & 13.59 & \underline{3.84} & 56.01 & 9.47 & 23.83 & 176.41 & 93.54 & 49.10 & 74.23\\
VGGT-Motion~\cite{vggtmotion}    & $\times$ & \textbf{6.79} & 83.29 & 66.48 & \underline{7.08} & \textbf{2.78} & \textbf{7.63} & \textbf{4.06} & 3.72 & \textbf{39.93} & \underline{28.77} & \textbf{15.35} & \underline{24.17}\\
VGGT-Long\textsuperscript{*}~\cite{vggtlong}        & $\times$ & 10.81 & \underline{66.46} & 56.00 & \textbf{5.56} & 3.93 & \underline{10.11} & \underline{6.90} & \textbf{2.83} & 57.45 & 31.17 & 22.11 & 24.85\\
SwiftVGGT\textsuperscript{*}~\cite{swiftvggt}       & $\times$ & \underline{9.92} & 114.96 & \underline{55.45} & 10.38 & 5.48 & 12.50 & 8.38 & 3.50 & 69.74 & 30.92 & 25.55 & 31.53\\
\rowcolor{gray!12}
\textbf{\methodname} (ours)      & $\times$ & 13.74 & \textbf{56.34} & \textbf{48.70} & 9.33 & 4.00 & 11.02 & 6.93 & \underline{3.15} & \underline{47.15} & \textbf{26.24} & \underline{22.05} & \textbf{22.61}\\
\bottomrule
\end{tabular}%
}
\end{table}
\Cref{tab:kitti_perseq} shows the results on KITTI Odometry. \methodname is particularly effective on long sequences where accumulated drift is severe. Compared with SwiftVGGT, ATE RMSE decreases from $114.96\,\mathrm{m}$ to $56.34\,\mathrm{m}$ on Seq.~01, from $69.74\,\mathrm{m}$ to $47.15\,\mathrm{m}$ on Seq.~08, and from $30.92\,\mathrm{m}$ to $26.24\,\mathrm{m}$ on Seq.~09, corresponding to reductions of 51.0\%, 32.4\%, and 15.1\%, respectively. \methodname also achieves the best result on Seq.~02, the longest sequence in terms of frame count. These results suggest that the proposed skip-edge formulation effectively reduces long-range drift in extended driving sequences. 
We additionally provide full-transformation results following the SwiftVGGT evaluation convention and cross-framework results on VGGT-Long in Supp.

\noindent\textbf{Waymo Open.}
\begin{table}[t]
\centering
\caption{\textbf{Performance comparison on Waymo Open.}
\textbf{Bold}/\underline{underline} mark the best/second-best among calibration-free methods.
``Calib.'' = uses camera intrinsics.
Results with * are reproduced by us.}
\label{tab:waymo_perseg}
\setlength{\tabcolsep}{2.5pt}
\renewcommand{\arraystretch}{1.1}
\scriptsize
\resizebox{\textwidth}{!}{%
\begin{tabular}{@{}l|c|rrrrrrrrr|r@{}}
\toprule
Method (segment ID prefix) & Calib. & 1634 & 1838 & 3156 & 3461 & 3711 & 4058 & 4604 & 5200 & 6104 & Avg. (\({\downarrow}\))\\
\midrule
\textit{\# frames} & & \textit{198} & \textit{199} & \textit{199} & \textit{199} & \textit{196} & \textit{199} & \textit{198} & \textit{199} & \textit{198} & \\
\textit{Length (m)} & & \textit{160} & \textit{42} & \textit{165} & \textit{351} & \textit{273} & \textit{86} & \textit{266} & \textit{135} & \textit{63} & \\
\textit{Speed (m/fr)} & & \textit{0.81} & \textit{0.21} & \textit{0.83} & \textit{1.77} & \textit{1.39} & \textit{0.43} & \textit{1.34} & \textit{0.68} & \textit{0.32} & \\
\textit{Traffic} & & \textit{Low} & \textit{High} & \textit{Low} & \textit{Low} & \textit{Med.} & \textit{Low} & \textit{Med.} & \textit{Low} & \textit{High} & \\
\midrule
DROID-SLAM~\cite{droidslam}      & \checkmark & 3.705 & 0.301 & 0.447 & 8.653 & 9.320 & 7.621 & 4.170 & TL & 0.264 & 4.396 \\
\midrule
MASt3R-SLAM~\cite{mast3rslam}    & $\times$ & 4.500 & 0.556 & 1.833 & 12.544 & 8.601 & 1.412 & 5.428 & 7.910 & 1.195 & 5.560 \\
CUT3R~\cite{cut3r}               & $\times$ & 8.781 & 3.810 & 5.790 & 24.015 & 13.070 & 7.261 & 13.206 & 8.597 & 3.229 & 9.872 \\
VGGT-SLAM~\cite{vggtslam}        & $\times$ & 1.820 & 0.640 & 1.420 & 12.020 & 7.440 & 2.140 & 2.950 & 7.990 & 0.570 & 4.110 \\
VGGT-Motion~\cite{vggtmotion}    & $\times$ & \textbf{1.346} & \textbf{0.139} & \underline{0.404} & 3.096 & \textbf{2.003} & \textbf{1.081} & 2.160 & 3.933 & \textbf{0.121} & 1.587 \\
VGGT-Long~\cite{vggtlong}        & $\times$ & \underline{1.753} & 2.629 & 0.559 & 3.452 & \underline{3.343} & 1.444 & \textbf{1.541} & 2.547 & 0.455 & 1.996 \\
SwiftVGGT\textsuperscript{*}~\cite{swiftvggt}       & $\times$ & 2.835 & \underline{0.166} & \textbf{0.221} & \underline{1.630} & 3.356 & 1.577 & 2.320 & \underline{1.785} & 0.125 & \underline{1.557}\\
\rowcolor{gray!12}
\textbf{\methodname} (ours)      & $\times$ & 2.412 & 0.183 & 0.540 & \textbf{0.698} & 3.918 & \underline{1.271} & \underline{2.034} & \textbf{1.428} & \underline{0.124} & \textbf{1.401}\\
\bottomrule
\end{tabular}%
}
\end{table}
\Cref{tab:waymo_perseg} reports per-segment results on nine short segments from the Waymo Open Dataset, each containing approximately 199 frames. \methodname achieves the best average ATE RMSE among all compared methods, reducing the error from $1.56\,\mathrm{m}$ for SwiftVGGT to $1.40\,\mathrm{m}$. It also outperforms VGGT-Motion and VGGT-Long, which obtain $1.59\,\mathrm{m}$ and $2.00\,\mathrm{m}$, respectively. These results show that the proposed skip-edge formulation remains effective even on short driving sequences with limited temporal context.

\noindent\textbf{Virtual KITTI.}
\Cref{tab:vkitti2_perseq} reports results on Virtual KITTI. Across five synthetic driving scenes and six appearance variants, \methodname achieves the best average ATE RMSE among calibration-free methods, with $2.03\,\mathrm{m}$ compared with $2.05\,\mathrm{m}$ for VGGT-Long and $2.50\,\mathrm{m}$ for SwiftVGGT. Together with the KITTI Odometry and Waymo results, this indicates that the skip-edge formulation in \methodname also generalizes to synthetic driving scenes with diverse appearance conditions.
\begin{table}[t]
\centering
\caption{\textbf{Performance comparison on Virtual KITTI.}
\textbf{Bold}/\underline{underline} mark the best/second-best among calibration-free methods.
``Calib.'' = uses camera intrinsics, ``TL'' = tracking lost, ``OOM'' = out of memory (on at least one weather variant within the scene).
Results with * are reproduced by us.
}
\label{tab:vkitti2_perseq}
\setlength{\tabcolsep}{1.8pt}
\renewcommand{\arraystretch}{1.0}
\scriptsize
\resizebox{\textwidth}{!}{%
\begin{tabular}{@{}l|c|rrrrr|r@{}}
\toprule
Method & Calib. & Scene01 & Scene02 & Scene06 & Scene18 & Scene20 & Avg. (\({\downarrow}\))\\
\midrule
DROID-SLAM~\cite{droidslam}      & \checkmark & 1.137 & 0.064 & 0.038 & 2.205 & 4.157 & 1.520 \\
\midrule
MASt3R-SLAM~\cite{mast3rslam}    & $\times$ & \multicolumn{5}{c|}{TL} & --\\
CUT3R~\cite{cut3r}               & $\times$ & 48.362 & 20.119 & 0.772 & 17.149 & 103.692 & 38.019 \\
Fast3R~\cite{fast3r}             & $\times$ & \multicolumn{5}{c|}{OOM} & --\\
VGGT-Long~\cite{vggtlong}        & $\times$ & \textbf{1.049} & 0.703 & 0.438 & \underline{1.397} & 6.683 & \underline{2.054} \\
SwiftVGGT\textsuperscript{*}~\cite{swiftvggt}       & $\times$ & 5.178 & \textbf{0.332} & \underline{0.186} & 1.817 & \underline{4.960} & 2.495 \\
\rowcolor{gray!12}
\textbf{\methodname} (ours)      & $\times$ & \underline{3.680} & \underline{0.335} & \textbf{0.170} & \textbf{1.268} & \textbf{4.681} & \textbf{2.027} \\
\bottomrule
\end{tabular}%
}
\end{table}

\section{Discussion}
\label{sec:discussion}
\subsection{Qualitative Results}

\vspace{-1.5em}
\begin{figure}[!htbp]
\centering
\includegraphics[width=\linewidth]{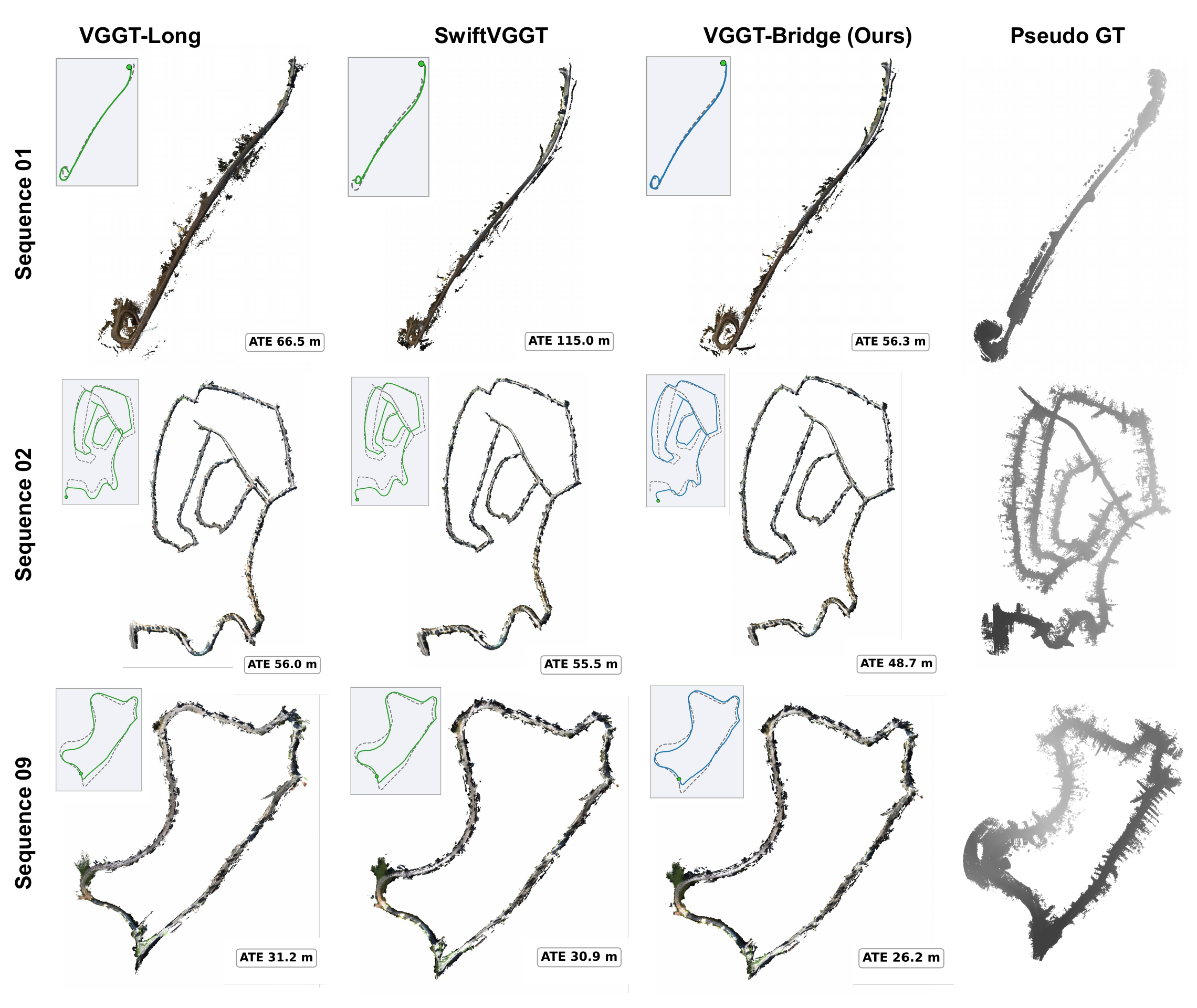}
\caption{\textbf{Camera trajectory and dense 3D reconstruction comparison on KITTI Odometry.} 
}
\label{fig:kitti_traj}

\end{figure}

\Cref{fig:kitti_traj} presents qualitative camera trajectory comparisons on representative sequences (Seq.~01, Seq.~02, Seq.~09) of the KITTI Odometry dataset.
Following SwiftVGGT~\cite{swiftvggt}, we build the pseudo ground-truth point cloud by accumulating LiDAR scans with the ground-truth poses.
Our method consistently achieves lower ATE than the baselines, and this improvement is also clearly visible in the trajectories.
The three sequences represent distinct loop conditions.
Seq.~01 is a loop-free highway sequence.
SwiftVGGT progressively underestimates the scale toward the end and its dense reconstruction collapses, whereas our method recovers the open-ended trajectory that matches the ground-truth.
On Seq.~02, where loop detection fails to close the loop precisely, our method still produces a trajectory better aligned with the ground-truth and lowers the ATE.
On Seq.~09, where the loop is correctly closed, the adaptive policy adds skip edges without conflicting with the loop closures and further improves the trajectory.
Overall, these qualitative results confirm that the skip edge and pseudo-loop closures induced by coarse chunks mitigate scale drift and yield globally consistent trajectories across different loop conditions, corroborating our quantitative results.
We further report the standard 3D reconstruction metrics of Accuracy, Completeness, and Chamfer Distance in Supp.

\subsection{Scale Bias Analysis and Ablations}
\label{sec:mechanism}

\begin{figure}[t]
\centering
\begin{minipage}[t]{0.6\linewidth}
\vspace{0pt}
\includegraphics[width=\linewidth]{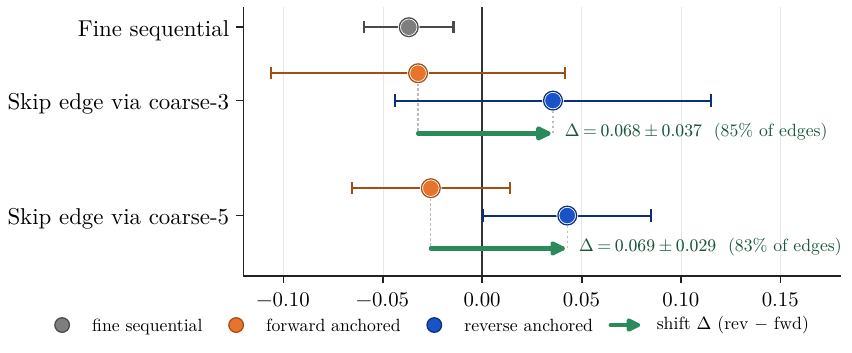}
\end{minipage}
\begin{minipage}[t]{0.36\linewidth}
\vspace{0pt}
\includegraphics[width=\linewidth]{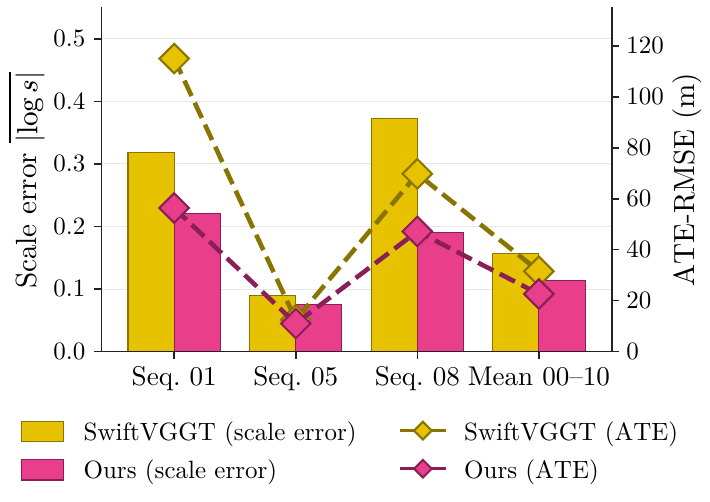}
\end{minipage}
\vspace{-0.5em}
\caption{\textbf{Scale bias on KITTI Odometry 00--10} ($n{=}11$ sequences).
\textbf{Left:} mean $\pm$ std of the pairwise Umeyama $\log s$ measured during chunk alignment, for sequential and coarse chunks.
The arrow marks the shift $\Delta$ from forward to reverse anchoring, with the percentage of edges where $\Delta{>}0$.
\textbf{Right:} residual local scale error $\overline{|\log s|}$ (bars) and the ATE RMSE (lines) of the final trajectory, for SwiftVGGT and our method.
The scale error is measured on sliding $75$-frame windows against the ground truth.}
\vspace{-1.0em}
\label{fig:signflip}
\end{figure}

We analyze VGGT's scale bias at two stages of the pipeline and confirm that reverse anchoring corrects it as intended.
We estimate the scale bias in chunk alignments and the residual scale error against the ground-truth, shown respectively in the left and right panels of \Cref{fig:signflip}.

\noindent\textbf{The bias exists and shifts with anchoring.}
We record the pairwise Umeyama scale $s$ of every alignment that forms an edge in the pose graph on KITTI Odometry 00--10, and report its mean and standard deviation over the eleven sequences.
Fine-sequential alignments are always forward-anchored, whereas each coarse chunk is measured both forward-anchored and reverse-anchored.
In the ideal case, aligning two chunks that reconstruct the same overlap region would give no scale difference, so $\log s = 0$.
The mean log scale across the fine-sequential edges is $-0.037$, meaning a forward-ordered chunk is reconstructed at roughly $3.6\%$ smaller scale than its predecessor, a one-sided shrinking bias.
Composed forward skip edges inherit this bias.
To isolate the effect of reversal, we compare forward- and reverse-anchored estimates for the same skip edges at each stride, and measure the per-edge shift as $\Delta{=}\log s_{\mathrm{rev}}{-}\log s_{\mathrm{fwd}}$.
Reverse anchoring moves the mean log scale from $-0.032$ to $+0.036$ at stride 3 and from $-0.026$ to $+0.043$ at stride 5.
$\Delta$ averages $0.068\pm0.037$ at stride 3 and $0.069\pm0.029$ at stride 5, and is positive for $85\%$ and $83\%$ of the edges, respectively.
The reverse skip edge therefore shifts the scale estimate in the direction opposite to the forward bias, supplying the anti-bias signal that the forward chain lacks.

\noindent\textbf{The anti-bias signal suppresses global scale drift.}
We apply the same log-scale analysis to the output trajectory against the ground truth.
For each sequence we first $\SimT$-align the whole predicted trajectory to the ground truth, which removes the global scale offset and leaves only local deviations to measure.
We then measure those deviations by sliding a $75$-frame window along the trajectory and fitting the $k$-th window to the ground truth with its own local $\SimT$ Umeyama, whose scale factor we record as $s^{\mathrm{local}}_k$.
Averaging $|\log s^{\mathrm{local}}_k|$ over all windows measures the residual local scale deviation after global optimization, and is trivially zero for the ground truth against itself.
Across KITTI Odometry 00--10, this deviation drops from $0.157$ for SwiftVGGT to $0.113$ for our method, a $28\%$ reduction toward the zero reference.
The largest gains fall on the drift-dominated sequences, from $0.319$ to $0.221$ on Seq.~01 and from $0.372$ to $0.190$ on Seq.~08.

\begin{table}[t]
\centering
\caption{\textbf{Component ablation on KITTI Odometry.}
Each row cumulatively adds one component on top of the previous.
Trans and Rot are the translation (m) and rotation ($^\circ$) ATE, and Scale is the mean $|\log s|$ of \Cref{fig:signflip}~(right).}
\label{tab:ablation}
\setlength{\tabcolsep}{2.5pt}
\renewcommand{\arraystretch}{1.1}
\scriptsize
\resizebox{\textwidth}{!}{%
\begin{tabular}{@{}l|rrrrrrrrrrr|r|rr@{}}
\toprule
 & \multicolumn{11}{c|}{Translation ATE (m)} & \multicolumn{3}{c}{Mean} \\
\cmidrule(lr){2-12}\cmidrule(l){13-15}
Configuration & 00 & 01 & 02 & 03 & 04 & 05 & 06 & 07 & 08 & 09 & 10 & Trans & Rot & Scale\\
\midrule
\textit{\# frames} & \textit{4541} & \textit{1101} & \textit{4661} & \textit{801} & \textit{271} & \textit{2761} & \textit{1101} & \textit{1101} & \textit{4071} & \textit{1591} & \textit{1201} & & & \\
\textit{Loop}     & \textit{\checkmark} & \textit{--} & \textit{\checkmark} & \textit{--} & \textit{--} & \textit{\checkmark} & \textit{\checkmark} & \textit{\checkmark} & \textit{--} & \textit{\checkmark} & \textit{--} & & & \\
\midrule
SwiftVGGT               &  9.92 & 114.96 & 55.45 & 10.38 & 5.48 & 12.50 & 8.38 & 3.50 & 69.74 & 30.92 & 25.55 & 31.53 & 13.38 & 0.157\\
+ coarse-stride chunk   & 13.00 &  69.68 & 53.43 &  9.26 & 4.44 & 12.30 & 7.28 & 4.68 & 72.43 & 24.06 & 25.45 & 26.91 & $\mathbf{12.15}$ & 0.139\\
+ reverse coarse chunk  & 15.01 &  56.35 & 49.80 &  9.50 & 3.97 & 10.88 & 8.87 & 3.16 & 45.53 & 27.59 & 20.74 & 22.85 & 12.41 & 0.115\\
\rowcolor{gray!12}
+ loop-aware (ours)     & 13.74 &  56.34 & 48.70 &  9.33 & 4.00 & 11.02 & 6.93 & 3.15 & 47.15 & 26.24 & 22.05 & $\mathbf{22.61}$ & 12.26 & $\mathbf{0.113}$\\
\bottomrule
\end{tabular}%
}
\end{table}

\begin{table*}[t]
\centering
\begin{minipage}[t]{0.42\textwidth}
\centering
\caption{\textbf{Anchoring position ablation on KITTI Odometry.}
ATE RMSE (m) per anchor position configuration is reported.}
\label{tab:dir_ablation}
\setlength{\tabcolsep}{4pt}
\scriptsize
\begin{tabular}{cccc}
\toprule
\# & Stride 3 & Stride 5 & Mean ATE\\
\midrule
(1) & fwd & fwd & 26.91\\
(2) & rev & rev & 24.08\\
\rowcolor{gray!12}
(3) & rev & fwd & $\mathbf{22.61}$\\
\bottomrule
\end{tabular}
\end{minipage}
\hfill
\begin{minipage}[t]{0.56\textwidth}
\centering
\caption{\textbf{Efficiency comparison on KITTI Odometry.}
Total processing time, FPS (relative to VGGT-Long), and mean ATE RMSE (m) over all sequences are reported.}
\label{tab:runtime}
\setlength{\tabcolsep}{4pt}
\resizebox{\textwidth}{!}{%
\scriptsize
\begin{tabular}{lccc}
\toprule
Method & Total time (s) & FPS & Mean ATE\\
\midrule
VGGT-Long~\cite{vggtlong} & 7{,}611 & 3.0 ($1.0\times$) & 24.85\\
SwiftVGGT~\cite{swiftvggt} & 2{,}583 & 9.0 ($2.9\times$) & 31.53\\
\rowcolor{gray!12}
\textbf{\methodname} & 3{,}883 & 6.0 ($2.0\times$) & 22.61\\
\bottomrule
\end{tabular}
}
\end{minipage}
\end{table*}

\noindent\textbf{Component ablation.}
We first conduct a component ablation, adding the three proposed components one at a time, as reported in \Cref{tab:ablation}.
To examine the contribution of each component more closely, we also report the mean rotation ATE and scale error.
Adding the coarse-stride skip edges lowers the mean ATE from $31.53$ to $26.91$ and reduces all three errors, with the largest gains on the long, drift-dominated sequences such as Seq.~01 and Seq.~09.
Reverse-anchoring the coarse chunks brings a further drop to $22.85$, driven by the drift correction it supplies where no loop closure is available, most visibly on Seq.~08.
It also brings the largest drop in scale error, from $0.139$ to $0.115$, without adding any new edge.
Finally, the loop-aware policy improves the mean to $22.61$ by restoring forward anchoring in loop-rich sequences such as Seq.~06, where a uniform reversal would conflict with the existing loop closures.
These results show that each component contributes to the final performance.

\noindent\textbf{Ablation on anchor position of coarse chunk.}
We next analyze the anchor position of the coarse chunks on a per-stride basis, as shown in \Cref{tab:dir_ablation}.
A chunk marked \emph{fwd} is kept forward-anchored, while a \emph{rev} chunk is anchored adaptively by the loop-aware policy.
Reverse-anchoring stride 3 while keeping stride 5 forward gives the best result of $22.61$.
Reverse-anchoring stride 5 as well degrades performance to $24.08$.
Although reversal shifts the scale at stride 5 as much as at stride 3, it also raises the Umeyama alignment residual of a stride-5 skip edge by $17\%$, against $10\%$ at stride 3, so the anti-bias signal comes with a less accurate alignment.
We therefore reverse-anchor stride 3 and keep stride 5 forward.

\subsection{Efficiency of \methodname}
\label{appendix:runtime}

To assess the efficiency of our method, we measure the total processing time and FPS on the KITTI Odometry benchmark using a single NVIDIA RTX 4090 GPU.
As shown in \Cref{tab:runtime}, compared to VGGT-Long, which relies on costly IRLS optimization, our method is $2.0\times$ faster while achieving $9.0\%$ lower ATE RMSE.
While $1.5\times$ slower than SwiftVGGT, which uses only sequential edges with Umeyama alignment, our method reduces ATE RMSE by $28.3\%$.
This indicates that augmenting the sequential edges between adjacent chunks with additional skip edges yields a substantial performance improvement.
Overall, our method achieves a better speed–accuracy trade-off than VGGT-Long, while improving the performance of SwiftVGGT at a reasonable additional cost.

\section{Conclusion}
\label{sec:conclusion}

We present \methodname, a training-free method that adds long-range skip edges to the chunk-and-align 3D reconstruction pipeline.
By running VGGT at coarse strides, each coarse chunk bridges non-adjacent fine chunks into a direct constraint.
Reverse anchoring then turns VGGT's first-frame scale bias into a drift correction, while a loop-aware policy keeps this reversal compatible with real loop closures.
Across KITTI Odometry, Waymo Open Dataset, and Virtual KITTI, \methodname attains the best mean ATE RMSE among all chunk-and-align methods, reducing the error by 28.3\%, 10.0\%, and 18.8\% over the SwiftVGGT baseline.
Our mechanism analysis confirms that reverse anchoring counteracts the scale bias of skip edges and suppresses the accumulated scale drift, and our ablations show that each component contributes to the final performance.
These results show that adding long-range skip edges is a practical and effective way to reduce accumulated drift in chunk-and-align reconstruction.

\section*{Acknowledgments}
This work was supported by {Institute of Information \& Communications Technology Planning \& Evaluation(IITP)} grant funded by the Korea government (MSIT) (RS-2024-00439020, Developing Sustainable, Real-Time Generative AI for Multimodal Interaction, SW Starlab).


\bibliographystyle{splncs04}
\bibliography{main}

\newpage
\section*{Appendix for \methodname{}: Beyond Sequential Pose Graphs via Coarse-Stride Skip Edges}
\appendix
\section{Implementation details}
\label{sec:appendix_impl}

\subsection{Hardware and software}
All experiments run on a single machine with an Intel Xeon Gold 6426Y CPU, $240$\,GB of system RAM, and NVIDIA GeForce RTX 4090 (24\,GB) GPUs.
Every reported runtime is measured on a single RTX 4090.
The software environment is Ubuntu~22.04 with CUDA~12.1, cuDNN~8.9, and PyTorch~2.3.1.

\subsection{Hyperparameters}
Table~\ref{tab:hyperparams} summarizes the hyperparameters used across datasets.
The fine chunk size $B$ and overlap $O$ follow the default settings of SwiftVGGT~\cite{swiftvggt}.
The coarse chunk size and overlap are fixed at $(B^{r}, O^{r}){=}(35,14)$ for stride~3 and $(25,10)$ for stride~5 across all datasets.
The skip-edge base weight is $\bar w{=}0.3$ on all benchmarks.
The adaptive policy threshold is $\tau{=}0.7$ with a loop-density window $W{=}60$, and the Cauchy kernel uses $\delta{=}2.0$.

\begin{table}[h]
\centering
\caption{\textbf{Hyperparameter settings.}}
\label{tab:hyperparams}
\begin{tabular}{llccc}
\toprule
\multicolumn{2}{l}{\textbf{Parameter}} & \textbf{KITTI Odometry} & \textbf{Virtual KITTI} & \textbf{Waymo Open} \\
\midrule
\multicolumn{2}{l}{Fine chunk size $B$}        & 75 & 75 & 60 \\
\multicolumn{2}{l}{Fine overlap $O$}           & 30 & 30 & 24 \\
\midrule
\multirow{2}{*}{Coarse ($r{=}3$)} & $B^{3}$    & 35 & 35 & 35 \\
                                  & $O^{3}$    & 14 & 14 & 14 \\
\multirow{2}{*}{Coarse ($r{=}5$)} & $B^{5}$    & 25 & 25 & 25 \\
                                  & $O^{5}$    & 10 & 10 & 10 \\
\midrule
\multicolumn{2}{l}{Skip-edge base weight $\bar w$} & 0.3 & 0.3 & 0.3 \\
\multicolumn{2}{l}{Policy threshold $\tau$ / window $W$} & 0.7 / 60 & 0.7 / 60 & 0.7 / 60 \\
\multicolumn{2}{l}{Cauchy kernel $\delta$} & 2.0 & 2.0 & 2.0 \\
\bottomrule
\end{tabular}
\end{table}

\section{Scale bias and hyperparameters across benchmarks}
\label{sec:appendix_bias}
We repeat the measurement of \Cref{fig:signflip}~(left) of the main paper on five benchmarks and run our method on each of them, as reported in \Cref{tab:appendix_bias}.
Besides the three driving benchmarks, we add two non-driving datasets, EuRoC MAV~\cite{euroc} and 7-Scenes~\cite{sevenscenes}.
EuRoC consists of eleven micro aerial vehicle sequences, and 7-Scenes consists of seven hand-held indoor scenes.
Both run with the same settings as KITTI Odometry.

The mean $\log s$ of the fine sequential edges is negative on all five benchmarks.
A forward-anchored chunk is thus reconstructed, on average, at a smaller scale than its predecessor, on hand-held and aerial motion as well as on driving.
This suggests that the anchoring bias is not specific to driving data.
Every choice of the skip-edge base weight $\bar w$ improves over SwiftVGGT on all five benchmarks.
The best value differs across benchmarks, as Virtual KITTI and EuRoC favor larger weights while KITTI Odometry and Waymo Open favor $\bar w{=}0.3$.
To avoid dataset-specific hyperparameter selection, we fix $\bar w{=}0.3$ on all benchmarks in the main paper, which still improves over SwiftVGGT on every benchmark.

\begin{table}[h]
\centering
\caption{\textbf{Anchoring bias and skip-edge weight across five benchmarks.}
Mean $\log s$ is the mean $\pm$ std of the fine sequential edges over the sequences, as in \Cref{fig:signflip}~(left) of the main paper.
The remaining columns report the mean translation ATE RMSE.
\textbf{Bold} marks the best result in each row.}
\label{tab:appendix_bias}
\resizebox{0.9\textwidth}{!}{%
\setlength{\tabcolsep}{3pt}
\begin{tabular}{l c c c ccc}
\toprule
 & & & & \multicolumn{3}{c}{\methodname{}} \\
\cmidrule(lr){5-7}
Dataset & \# seq. & Mean $\log s$ & SwiftVGGT & $\bar w{=}0.3$ & $\bar w{=}0.7$ & $\bar w{=}1.0$ \\
\midrule
KITTI Odometry & 11 & $-0.0370\pm0.0225$ & 31.53 & \textbf{22.61} & 22.82 & 23.06 \\
Virtual KITTI  & 30 & $-0.0761\pm0.0896$ & 2.495 & 2.027 & 1.853 & \textbf{1.790} \\
Waymo Open     & 9  & $-0.0470\pm0.0341$ & 1.557 & \textbf{1.401} & 1.406 & 1.408 \\
\midrule
EuRoC MAV      & 11 & $-0.0387\pm0.0221$ & 1.722 & 1.698 & \textbf{1.632} & 1.638 \\
7-Scenes       & 7  & $-0.0167\pm0.0169$ & 0.087 & 0.083 & \textbf{0.082} & \textbf{0.082} \\
\bottomrule
\end{tabular}%
}
\end{table}

\section{Cross-framework transfer: VGGT-Long}
\label{sec:appendix_vggtlong}
To test whether our contribution is tied to a particular alignment style or is a general pose-graph augmentation, we transplant the full \methodname{} recipe onto VGGT-Long~\cite{vggtlong}.
The two backbones differ only in how each chunk alignment $\SimT$ is estimated, while sharing the same $\SimT$ pose graph and the same Levenberg--Marquardt~\cite{levenberg, marquardt} solver.
SwiftVGGT estimates it with a single Umeyama step on the filtered 3D points, as in our default setting.
VGGT-Long instead runs Iteratively Reweighted Least Squares (IRLS), repeating a weighted Umeyama step while recomputing the Huber term of each weight from its current residual.
Each weight combines the VGGT confidence with a Huber term that downweights high-residual points, so unreliable ones contribute less as the iterations proceed.
The two estimators differ sharply in this respect, so transferring the same method across them tests whether the skip edge is specific to the single-step Umeyama estimator or generalizes to an iteratively reweighted one.


The three method components transfer as is, namely the coarse-stride skip edge, the reverse-anchored coarse chunk, and the loop-aware adaptive direction policy.
The only change is inside skip-edge composition, where each fine--coarse $\SimT$ is now solved by the confidence-weighted IRLS estimator rather than a single Umeyama step, matching the host backbone.
In \Cref{eq:pgo_full}, the edge type with the larger alignment residual can dominate the optimization, so we set the skip-edge scalars from the ratio of skip-edge to sequential-edge residuals.
On SwiftVGGT, the median Umeyama residual of a skip edge is $0.106$ against $0.052$ for a sequential edge, a ratio of $2.0\times$.
IRLS on VGGT-Long tightens the sequential edges to $0.024$ while the skip edges stay at $0.123$, raising the ratio to $5.1\times$.
We therefore recalibrate three scalars so the looser skip constraint is weighted appropriately against the tighter sequential and loop edges.
We tighten the Huber threshold to $\delta{=}0.05$, so the IRLS reweighting discounts high-residual correspondences more aggressively.
We lower the skip-edge base weight to $\bar w{=}0.1$, so that the looser skip edges do not dominate the tighter sequential edges.
We set the loop-aware adaptive direction threshold to $\tau{=}0.05$, so reverse anchoring is applied only where loops are essentially absent.

\Cref{tab:appendix_vggtlong} reports the transfer alongside our main single-step-Umeyama result.
The skip-edge construction also proves effective on the IRLS backbone of VGGT-Long.
As on the Umeyama backbone, the gains concentrate on the long, drift-dominated sequences, where attaching our method lowers the ATE RMSE by $37.6\%$ on Seq.~01 and $17.1\%$ on Seq.~08.
A few loop-rich sequences (00, 06, 07) show small regressions, at most $1.5$\,m, but these are outweighed by the larger gains, giving a net $17.1\%$ reduction in the mean.
The same coarse-stride skip edge construction improves two structurally different alignment estimators, suggesting that the skip edge is not tied to the single-step Umeyama estimator and can potentially extend to other chunk-and-align backbones.

\begin{table}[h]
\centering
\caption{\textbf{Cross-framework transfer of \methodname{} on VGGT-Long.} (per-sequence translation ATE RMSE (m) on KITTI Odometry).
The same skip-edge construction is added to two backbones, and $\Delta$ is the change relative to each backbone.}
\label{tab:appendix_vggtlong}
\setlength{\tabcolsep}{2.5pt}
\scriptsize
\resizebox{\textwidth}{!}{%
\begin{tabular}{l rrrrr rrrrr r r}
\toprule
Method & 00 & 01 & 02 & 03 & 04 & 05 & 06 & 07 & 08 & 09 & 10 & Mean\\
\midrule
SwiftVGGT~\cite{swiftvggt} & 9.92 & 114.96 & 55.45 & 10.38 & 5.48 & 12.50 & 8.38 & 3.50 & 69.74 & 30.92 & 25.55 & 31.53\\
\textbf{\methodname} & 13.74 & 56.34 & 48.70 & 9.33 & 4.00 & 11.02 & 6.93 & 3.15 & 47.15 & 26.24 & 22.05 & \textbf{22.61}\\
\quad $\Delta$ (\%) & $+38.5$ & $-51.0$ & $-12.2$ & $-10.1$ & $-27.0$ & $-11.8$ & $-17.3$ & $-10.0$ & $-32.4$ & $-15.1$ & $-13.7$ & $\mathbf{-28.3}$\\
\midrule
VGGT-Long~\cite{vggtlong} & 10.81 & 66.46 & 56.00 & 5.56 & 3.93 & 10.11 & 6.90 & 2.83 & 57.45 & 31.17 & 22.11 & 24.85\\
\quad\textbf{+ \methodname} & 12.08 & 41.47 & 49.21 & 6.57 & 3.20 & 10.15 & 7.65 & 4.30 & 47.61 & 23.92 & 20.50 & \textbf{20.61}\\
\quad $\Delta$ (\%) & $+11.7$ & $-37.6$ & $-12.1$ & $+18.2$ & $-18.6$ & $+0.4$ & $+10.9$ & $+51.9$ & $-17.1$ & $-23.3$ & $-7.3$ & $\mathbf{-17.1}$\\
\bottomrule
\end{tabular}%
}
\end{table}

\section{Chamfer Evaluation}
\label{sec:appendix_evalrecon}
Beyond camera tracking, we evaluate the quality of the reconstructed 3D geometry against the LiDAR ground truth on KITTI Odometry and Waymo Open.
Following prior work~\cite{vggtlong,swiftvggt}, we build the ground truth by accumulating LiDAR scans with the ground-truth poses.
The KITTI Odometry ground truth is the accumulated LiDAR scan with a front-facing crop.
On KITTI Odometry, we evaluate the ten sequences 00--02 and 04--10, omitting Seq.~03 whose raw LiDAR scan is not part of the released KITTI raw data.
The Waymo Open ground truth is the LiDAR scan cropped to the front three camera views.

For each sequence, we merge the reconstructed point maps of each frame into a single global point cloud.
The point clouds of all methods are randomly downsampled to a common density, coarsely aligned to the ground truth with a $\SimT$ Umeyama step, and refined with Iterative Closest Point (ICP).
Under the same pre-processing protocol, we report three metrics.
Accuracy (Acc) is the RMSE of nearest-neighbor distances from the prediction to the ground truth, Completeness (Comp) is the RMSE in the reverse direction, and the Chamfer Distance (CD) is the mean of the two.
Lower is better for all three.


\begin{table}[t]
\centering
\caption{\textbf{Reconstruction quality on KITTI Odometry.} Acc / Comp / CD (m), lower is better. Averaged over the ten sequences with available raw LiDAR scan, omitting Seq.~03. Best per metric per sequence in \textbf{bold}.}
\label{tab:chamfer_kitti}
\resizebox{0.9\textwidth}{!}{%
\setlength{\tabcolsep}{3pt}
\scriptsize
\begin{tabular}{l ccc ccc ccc}
\toprule
 & \multicolumn{3}{c}{VGGT-Long~\cite{vggtlong}} & \multicolumn{3}{c}{SwiftVGGT~\cite{swiftvggt}} & \multicolumn{3}{c}{\methodname (ours)} \\
\cmidrule(lr){2-4}\cmidrule(lr){5-7}\cmidrule(lr){8-10}
Seq & Acc & Comp & CD & Acc & Comp & CD & Acc & Comp & CD \\
\midrule
00 & \textbf{1.62} & 7.31 & \textbf{4.47} & 1.95 & \textbf{7.24} & 4.60 & 2.11 & 8.78 & 5.45 \\
01 & 42.73 & 16.83 & 29.78 & 34.65 & 20.09 & 27.37 & \textbf{20.91} & \textbf{7.86} & \textbf{14.38} \\
02 & 18.72 & 30.28 & 24.50 & 17.82 & 32.57 & 25.19 & \textbf{13.72} & \textbf{29.19} & \textbf{21.45} \\
04 & \textbf{0.80} & 6.80 & \textbf{3.80} & 1.59 & \textbf{6.53} & 4.06 & 1.59 & 6.84 & 4.21 \\
05 & \textbf{1.24} & \textbf{7.17} & \textbf{4.21} & 4.38 & 9.29 & 6.84 & 1.73 & 7.55 & 4.64 \\
06 & 1.60 & \textbf{3.91} & \textbf{2.76} & 1.94 & 4.99 & 3.47 & \textbf{1.37} & 4.80 & 3.08 \\
07 & \textbf{0.79} & \textbf{6.37} & \textbf{3.58} & 1.10 & 6.95 & 4.02 & 1.03 & 6.67 & 3.85 \\
08 & \textbf{10.76} & 33.52 & 22.14 & 22.15 & 48.29 & 35.22 & 14.79 & \textbf{28.25} & \textbf{21.52} \\
09 & 5.08 & 15.29 & 10.18 & 4.76 & 15.67 & 10.22 & \textbf{3.96} & \textbf{13.81} & \textbf{8.88} \\
10 & \textbf{1.93} & \textbf{13.22} & \textbf{7.57} & 7.79 & 16.62 & 12.21 & 4.40 & \textbf{13.22} & 8.81 \\
\midrule
Mean & 8.53 & 14.07 & 11.30 & 9.81 & 16.83 & 13.32 & \textbf{6.56} & \textbf{12.70} & \textbf{9.63} \\
\bottomrule
\end{tabular}%
}
\end{table}

\begin{table}[t]
\centering
\caption{\textbf{Reconstruction quality on Waymo Open.} Acc / Comp / CD (m), lower is better. Best per metric per segment in \textbf{bold}.}
\label{tab:chamfer_waymo}
\resizebox{0.9\textwidth}{!}{%
\setlength{\tabcolsep}{3pt}
\scriptsize
\begin{tabular}{l ccc ccc ccc}
\toprule
 & \multicolumn{3}{c}{VGGT-Long~\cite{vggtlong}} & \multicolumn{3}{c}{SwiftVGGT~\cite{swiftvggt}} & \multicolumn{3}{c}{\methodname (ours)} \\
\cmidrule(lr){2-4}\cmidrule(lr){5-7}\cmidrule(lr){8-10}
Segment & Acc & Comp & CD & Acc & Comp & CD & Acc & Comp & CD \\
\midrule
1634 & \textbf{0.87} & 2.11 & 1.49 & 1.02 & \textbf{1.81} & \textbf{1.42} & 0.94 & 1.90 & \textbf{1.42} \\
1838 & \textbf{0.45} & 5.87 & \textbf{3.16} & 1.38 & 5.85 & 3.62 & 1.35 & \textbf{5.44} & 3.39 \\
3156 & \textbf{0.66} & 2.58 & 1.62 & 1.00 & 2.26 & 1.63 & 0.99 & \textbf{2.11} & \textbf{1.55} \\
3461 & \textbf{0.93} & \textbf{4.47} & \textbf{2.70} & 1.38 & 5.53 & 3.45 & 1.36 & 5.50 & 3.43 \\
3711 & \textbf{1.96} & 4.93 & 3.44 & 2.60 & \textbf{4.26} & 3.43 & 2.44 & 4.37 & \textbf{3.41} \\
4058 & 0.46 & \textbf{3.26} & \textbf{1.86} & 0.44 & 4.04 & 2.24 & \textbf{0.39} & 4.09 & 2.24 \\
4604 & \textbf{1.00} & 3.40 & \textbf{2.20} & 1.85 & \textbf{3.05} & 2.45 & 1.78 & 3.07 & 2.43 \\
5200 & \textbf{1.05} & 4.12 & \textbf{2.59} & 1.59 & \textbf{3.71} & 2.65 & 1.54 & 3.74 & 2.64 \\
6104 & 1.51 & \textbf{2.19} & 1.85 & 1.11 & 2.40 & 1.76 & \textbf{1.10} & 2.35 & \textbf{1.73} \\
\midrule
Mean & \textbf{0.99} & 3.66 & \textbf{2.32} & 1.38 & 3.66 & 2.52 & 1.32 & \textbf{3.62} & 2.47 \\
\bottomrule
\end{tabular}%
}
\end{table}

As shown in \Cref{tab:chamfer_kitti}, on KITTI Odometry benchmark, \methodname{} attains the best mean on all three metrics, with the lowest Accuracy of $6.56$, Completeness of $12.70$, and Chamfer Distance of $9.63$ among the compared methods.
These reconstruction results mirror the trajectory-accuracy trends.
The gains concentrate on the long drift-dominated sequences 01, 02, 08, and 09, where suppressing the accumulated drift also sharpens the reconstructed geometry.
On Seq.~01, the Chamfer Distance drops to $14.38$, less than half of $29.78$ for VGGT-Long and $27.37$ for SwiftVGGT.
On Seq.~08, it drops to $21.52$ from $35.22$ for SwiftVGGT, while remaining close to the $22.14$ of VGGT-Long.
On Waymo, which consists of short segments with little accumulated drift, \Cref{tab:chamfer_waymo} shows that \methodname{} attains the best mean Completeness of $3.62$.
Overall, these results show that \methodname{} improves not only the camera trajectory but also the quality of the reconstructed 3D geometry, and that the two gains share the same source, namely the suppression of long-range drift.

\begin{table}[h]
\centering
\caption{\textbf{Per-scene, per-condition ATE RMSE (m) on Virtual KITTI.}
Best among the calibration-free methods in \textbf{bold}.
``$\dagger$'' uses camera intrinsics, all other methods are calibration-free.
``TL'' = tracking lost, ``OOM'' = out of memory.
Results with * are reproduced by us.}
\label{tab:appendix_vkitti2_perscene}
\setlength{\tabcolsep}{2.5pt}
\scriptsize
\resizebox{\textwidth}{!}{%
\begin{tabular}{l|cccccc|cccccc}
\toprule
& \multicolumn{6}{c|}{\textbf{Scene~01}} & \multicolumn{6}{c}{\textbf{Scene~02}}\\
Method & {\tiny Clone} & {\tiny Fog} & {\tiny Morning} & {\tiny Overcast} & {\tiny Rain} & {\tiny Sunset} & {\tiny Clone} & {\tiny Fog} & {\tiny Morning} & {\tiny Overcast} & {\tiny Rain} & {\tiny Sunset}\\
\midrule
DROID-SLAM$^{\dagger}$~\cite{droidslam} & 1.03 & 1.87 & 0.99 & 1.01 & 0.78 & 1.15 & 0.10 & 0.04 & 0.05 & 0.05 & 0.04 & 0.11\\
MASt3R-SLAM~\cite{mast3rslam} & \multicolumn{12}{c}{TL}\\
CUT3R~\cite{cut3r} & 43.30 & 62.19 & 50.61 & 38.73 & 51.55 & 43.79 & 23.77 & 9.95 & 28.41 & 24.64 & 7.96 & 25.97\\
Fast3R~\cite{fast3r} & \multicolumn{12}{c}{OOM}\\
VGGT-Long~\cite{vggtlong} & \textbf{0.76} & 0.87 & \textbf{0.93} & 0.67 & \textbf{1.80} & \textbf{1.26} & 0.72 & 0.71 & 0.72 & 0.68 & 0.69 & 0.69\\
SwiftVGGT\textsuperscript{*}~\cite{swiftvggt} & 4.13 & 0.55 & 16.29 & 1.03 & 6.94 & 2.12 & 0.43 & 0.32 & 0.37 & \textbf{0.26} & \textbf{0.34} & 0.27\\
\textbf{\methodname} & 2.90 & \textbf{0.54} & 10.76 & \textbf{0.64} & 5.64 & 1.59 & \textbf{0.38} & \textbf{0.30} & \textbf{0.37} & 0.26 & 0.44 & \textbf{0.25}\\
\midrule
& \multicolumn{6}{c|}{\textbf{Scene~06}} & \multicolumn{6}{c}{\textbf{Scene~18}}\\
Method & {\tiny Clone} & {\tiny Fog} & {\tiny Morning} & {\tiny Overcast} & {\tiny Rain} & {\tiny Sunset} & {\tiny Clone} & {\tiny Fog} & {\tiny Morning} & {\tiny Overcast} & {\tiny Rain} & {\tiny Sunset}\\
\midrule
DROID-SLAM$^{\dagger}$~\cite{droidslam} & 0.06 & 0.02 & 0.03 & 0.05 & TL & 0.02 & 2.48 & 2.03 & 1.89 & 2.33 & 2.55 & 1.94\\
MASt3R-SLAM~\cite{mast3rslam} & \multicolumn{12}{c}{TL}\\
CUT3R~\cite{cut3r} & 0.84 & 0.41 & 0.60 & 0.72 & 1.06 & 1.01 & 19.44 & 8.63 & 6.72 & 20.21 & 16.78 & 31.12\\
Fast3R~\cite{fast3r} & \multicolumn{12}{c}{OOM}\\
VGGT-Long~\cite{vggtlong} & 0.36 & 0.54 & 0.38 & 0.40 & 0.56 & 0.38 & \textbf{1.65} & 0.80 & 1.29 & \textbf{1.26} & 1.65 & 1.74\\
SwiftVGGT\textsuperscript{*}~\cite{swiftvggt} & 0.13 & 0.27 & 0.13 & 0.13 & 0.30 & 0.16 & 2.25 & 0.67 & 1.40 & 2.06 & 2.08 & 2.45\\
\textbf{\methodname} & \textbf{0.12} & \textbf{0.24} & \textbf{0.11} & \textbf{0.13} & \textbf{0.27} & \textbf{0.15} & 1.71 & \textbf{0.63} & \textbf{1.01} & 1.44 & \textbf{1.35} & \textbf{1.47}\\
\midrule
& \multicolumn{6}{c|}{\textbf{Scene~20}} & \multicolumn{6}{c}{\textbf{Average}}\\
Method & {\tiny Clone} & {\tiny Fog} & {\tiny Morning} & {\tiny Overcast} & {\tiny Rain} & {\tiny Sunset} & 01 & 02 & 06 & 18 & 20 & All\\
\midrule
DROID-SLAM$^{\dagger}$~\cite{droidslam} & 3.59 & 5.08 & 3.73 & 3.85 & 3.78 & 4.91 & 1.14 & 0.06 & 0.04 & 2.20 & 4.16 & 1.52\\
MASt3R-SLAM~\cite{mast3rslam} & \multicolumn{12}{c}{TL}\\
CUT3R~\cite{cut3r} & 129.50 & 76.96 & 117.95 & 114.51 & 66.70 & 116.53 & 48.36 & 20.12 & 0.77 & 17.15 & 103.69 & 38.02\\
Fast3R~\cite{fast3r} & \multicolumn{12}{c}{OOM}\\
VGGT-Long~\cite{vggtlong} & 9.66 & 8.19 & 6.34 & 4.56 & 6.50 & 4.85 & \textbf{1.05} & 0.70 & 0.44 & 1.40 & 6.68 & 2.05\\
SwiftVGGT\textsuperscript{*}~\cite{swiftvggt} & 7.00 & 8.61 & \textbf{4.27} & \textbf{4.48} & \textbf{3.01} & 2.38 & 5.18 & \textbf{0.33} & 0.19 & 1.82 & 4.96 & 2.50\\
\textbf{\methodname} & \textbf{5.82} & \textbf{5.88} & 5.06 & 4.89 & 4.33 & \textbf{2.09} & 3.68 & 0.33 & \textbf{0.17} & \textbf{1.27} & \textbf{4.68} & \textbf{2.03}\\
\bottomrule
\end{tabular}%
}
\end{table}

\section{Virtual KITTI: per-scene breakdown}
\label{sec:appendix_vkitti2_perscene}

\Cref{tab:appendix_vkitti2_perscene} expands the per-scene means of \Cref{tab:vkitti2_perseq} into a full table over all 30 sequences, covering five scenes under six weather conditions each.
\methodname{} attains the best average among the calibration-free methods and stays robust across the six weather conditions.

\section{Per-component runtime breakdown and memory}
\label{appendix:runtime_breakdown}
To isolate where the $1.5\times$ slowdown in \Cref{tab:runtime} of the main paper comes from, we measure per-stage wall-clock time on a short and a long sequence, KITTI Odometry Seq.~07 (1{,}101 frames) and Seq.~08 (4{,}071 frames).
While \Cref{tab:runtime} times the full pipeline, here we report only the major computation stages, so that the effect of adding \methodname{} is visible per stage.
\Cref{tab:runtime_breakdown} reports the breakdown together with the coarse chunk and skip edge counts that drive each stage.
The added cost is concentrated in the coarse VGGT~\cite{vggt} inference and the skip-edge composition.
These two stages account for nearly all of the overhead, adding $+26.3$\,s and $+28.5$\,s on Seq.~07 and $+110.3$\,s and $+38.7$\,s on Seq.~08.
Each coarse chunk spans far fewer frames than a fine chunk, with $B^{3}{=}35$ and $B^{5}{=}25$ against the $75$ frames of a fine chunk.
The extra VGGT inference therefore stays a moderate trade-off, despite the larger number of coarse chunks.
The pose graph optimization absorbs the additional skip edges with little overhead, since skip edges add only constraints and the number of pose variables to solve stays the same.
Even on Seq.~08 with $98$ skip edges, its solve time rises only from $0.1$\,s to $1.0$\,s.
The remaining stages take almost the same time in both methods.

\Cref{tab:appendix_memory} reports the peak memory on KITTI Odometry 00--10.
The peak GPU memory of our method is identical to SwiftVGGT on every sequence, since each coarse chunk feeds fewer frames to VGGT than a fine chunk.
The host memory grows, because the coarse chunk predictions are kept until the skip edges are formed.

\begin{table}[h]
\centering
\caption{\textbf{Per-stage runtime breakdown on KITTI Odometry Seq.~07 and Seq.~08.}
Seq.~07 has one detected loop pair and activates the loop pipeline, while Seq.~08 has none detected.
The loop pipeline is identical in both methods, so only the stages \methodname{} modifies differ.}
\label{tab:runtime_breakdown}
\resizebox{\textwidth}{!}{%
\setlength{\tabcolsep}{4pt}
\scriptsize
\begin{tabular}{l rr r rr r}
\toprule
 & \multicolumn{3}{c}{\textbf{KITTI Seq.~07} (1{,}101 frames)} & \multicolumn{3}{c}{\textbf{KITTI Seq.~08} (4{,}071 frames)} \\
\cmidrule(lr){2-4}\cmidrule(lr){5-7}
Stage & SwiftVGGT & Ours & $\Delta$ & SwiftVGGT & Ours & $\Delta$ \\
\midrule
Loop detection &    3.1 &    3.0 & $ -0.1$ &   10.2 &   10.1 & $ -0.1$ \\
VGGT chunk inference &   58.1 &   84.4 & $+26.3$ &  197.6 &  307.9 & $+110.3$ \\
Loop-centric chunk inference &    1.1 &    1.3 & $ +0.2$ &    0.0 &    0.0 & $ +0.0$ \\
Loop-closure alignment &    0.9 &    1.1 & $ +0.2$ &    0.0 &    0.0 & $ +0.0$ \\
Sequential-edge composition &   29.9 &   30.0 & $ +0.1$ &   76.0 &   76.0 & $ +0.0$ \\
Skip-edge composition &    0.0 &   28.5 & $+28.5$ &    0.0 &   38.7 & $+38.7$ \\
Pose graph optimization &    0.3 &    0.5 & $ +0.2$ &    0.1 &    1.0 & $ +0.9$ \\
\midrule
\textbf{Total} & \textbf{ 93.4} & \textbf{148.8} & $\mathbf{+55.4}$ & \textbf{283.9} & \textbf{433.7} & $\mathbf{+149.8}$ \\
\midrule
\multicolumn{7}{l}{\emph{Counts driving each stage}}\\
Fine chunks (stride 1) & 24 & 24 & --- & 90 & 90 & --- \\
Loop pairs & 1 & 1 & --- & 0 & 0 & --- \\
Coarse chunks (stride 3) & 0 & 17 & --- & 0 & 64 & --- \\
Coarse chunks (stride 5) & 0 & 15 & --- & 0 & 54 & --- \\
Skip edges & 0 & 26 & --- & 0 & 98 & --- \\
\bottomrule
\end{tabular}
}
\end{table}

\begin{table}[h]
\centering
\caption{\textbf{Peak memory on KITTI Odometry 00--10} (GiB), as the mean and the maximum over the eleven sequences.
GPU is the peak memory reserved by PyTorch, and host is the peak resident memory.}
\label{tab:appendix_memory}
\setlength{\tabcolsep}{4pt}
\scriptsize
\begin{tabular}{lcccc}
\toprule
 & \multicolumn{2}{c}{GPU} & \multicolumn{2}{c}{Host} \\
\cmidrule(lr){2-3}\cmidrule(lr){4-5}
Method & Mean & Max & Mean & Max \\
\midrule
SwiftVGGT~\cite{swiftvggt} & 10.30 & 14.49 & 15.74 & 29.90 \\
\methodname{} & 10.30 & 14.49 & 19.81 & 38.90 \\
\bottomrule
\end{tabular}
\end{table}

\section{Compute-matched comparison}
\label{sec:appendix_compute}
Our method feeds more frames to VGGT than SwiftVGGT, since it runs the coarse chunks in addition to the fine chunks.
To test whether the gain comes from this extra computation alone, we add computation to the default SwiftVGGT setting ($B{=}75$, $O{=}30$) in two ways, as reported in \Cref{tab:appendix_compute}.
The first raises the overlap $O$, giving each sequential alignment more shared frames.
The second runs an additional tiling with chunk size $B'{=}90$ and jointly optimizes both tilings in one pose graph.
Both variants feed VGGT as many frames as our method or more, yet both worsen the mean ATE of SwiftVGGT, while our method lowers it to $22.61$.
The gain therefore comes from our design rather than from the extra computation.

\begin{table}[h]
\centering
\caption{\textbf{Compute-matched comparison on KITTI Odometry 00--10.}
Frames to VGGT is the total number of frames fed to VGGT, relative to SwiftVGGT.}
\label{tab:appendix_compute}
\setlength{\tabcolsep}{4pt}
\scriptsize
\begin{tabular}{lccccc}
\toprule
 & \multicolumn{3}{c}{SwiftVGGT, $O$} & SwiftVGGT & \methodname{} \\
\cmidrule(lr){2-4}
 & $30$ & $45$ & $60$ & $+B'{=}90$ & $O{=}30$ \\
\midrule
Frames to VGGT & $1.00\times$ & $1.49\times$ & $2.94\times$ & $2.00\times$ & $1.53\times$ \\
Mean ATE (m) & 31.53 & 32.61 & 34.70 & 32.91 & \textbf{22.61} \\
\bottomrule
\end{tabular}
\end{table}

\section{Full-transformation evaluation}
\label{sec:appendix_fulltransform}
The main paper reports the translation-only ATE RMSE following VGGT-Long.
Here we additionally report the full-transformation ATE used by SwiftVGGT, which scores the full $\SimT$ pose rather than the translation alone.
Both conventions apply the same global alignment before scoring and differ only in how each residual pose is measured.
The translation-only ATE counts only the translational part of the residual, whereas the full-transformation ATE also includes the rotational part, so it additionally reflects orientation error.

\Cref{tab:fulltransform,tab:fulltransform_waymo} report the full-transformation ATE on KITTI Odometry and Waymo Open, comparing \methodname{} against VGGT-Long and SwiftVGGT.
\methodname{} achieves the lowest mean full-transformation ATE on both datasets, showing that its strong performance holds robustly across both metrics.
We report the translation-only convention in the main tables because most baselines use it, and we include the full-transformation results for direct comparison with SwiftVGGT.

\begin{table}[h]
\centering
\caption{\textbf{Full-transformation ATE on KITTI Odometry.}
\textbf{Bold} marks the best result.}
\label{tab:fulltransform}
\setlength{\tabcolsep}{2.5pt}
\renewcommand{\arraystretch}{1.1}
\scriptsize
\resizebox{\textwidth}{!}{%
\begin{tabular}{@{}l|rrrrrrrrrrr|r@{}}
\toprule
Method & 00 & 01 & 02 & 03 & 04 & 05 & 06 & 07 & 08 & 09 & 10 & Avg. (\({\downarrow}\))\\
\midrule
VGGT-Long~\cite{vggtlong} & 9.87 & 111.06 & 37.56 & \textbf{4.89} & \textbf{3.75} & \textbf{9.09} & \textbf{7.47} & \textbf{4.02} & 62.86 & 47.48 & \textbf{25.49} & 29.41\\
SwiftVGGT~\cite{swiftvggt} & \textbf{8.17} & 102.53 & \textbf{36.49} & 8.12 & 4.88 & 11.94 & 8.88 & 5.01 & 64.68 & 44.13 & 26.18 & 29.18\\
\rowcolor{gray!12}
\textbf{\methodname} (ours) & 11.61 & \textbf{61.21} & 38.18 & 7.50 & 3.91 & 9.92 & 7.71 & 4.63 & \textbf{49.29} & \textbf{35.19} & 25.84 & \textbf{23.18}\\
\bottomrule
\end{tabular}%
}
\end{table}

\begin{table}[h]
\centering
\caption{\textbf{Full-transformation ATE on Waymo Open.}
\textbf{Bold} marks the best result.}
\label{tab:fulltransform_waymo}
\setlength{\tabcolsep}{2.5pt}
\renewcommand{\arraystretch}{1.1}
\scriptsize
\resizebox{\textwidth}{!}{%
\begin{tabular}{@{}l|rrrrrrrrr|r@{}}
\toprule
Method (segment ID prefix) & 1634 & 1838 & 3156 & 3461 & 3711 & 4058 & 4604 & 5200 & 6104 & Avg. (\({\downarrow}\))\\
\midrule
VGGT-Long~\cite{vggtlong} & \textbf{3.09} & 2.54 & \textbf{2.34} & 4.01 & \textbf{4.04} & 3.14 & 2.87 & 3.41 & 2.33 & 3.08\\
SwiftVGGT~\cite{swiftvggt} & 3.11 & 2.45 & 2.35 & 2.72 & 4.08 & 3.11 & \textbf{2.84} & 2.83 & \textbf{2.21} & 2.85\\
\rowcolor{gray!12}
\textbf{\methodname} (ours) & 3.20 & \textbf{2.01} & 2.36 & \textbf{2.28} & 4.53 & \textbf{2.94} & 3.03 & \textbf{2.66} & 2.22 & \textbf{2.80}\\
\bottomrule
\end{tabular}%
}
\end{table}

\section{Additional hyperparameter ablations}
\label{sec:appendix_hpablation}
We sweep three design choices of \methodname, the coarse stride set $\mathcal{R}$, the loop-density window $W$, and the adaptive direction threshold $\tau$, across all 11 KITTI Odometry sequences.
All other hyperparameters are kept at their defaults.
Results are summarized in \Cref{tab:appendix_hpablation}.

\begin{table}[t]
\centering
\caption{\textbf{Hyperparameter sweeps on KITTI Odometry 00--10.}
$\Delta$ is the change relative to the default \methodname configuration ($\mathcal{R}{=}\{3,5\}$, $W{=}60$, $\tau{=}0.7$).
Translation-only ATE RMSE (m) reported, as in the main tables.}
\label{tab:appendix_hpablation}
\setlength{\tabcolsep}{8pt}
\small
\begin{tabular}{l c c}
\toprule
Configuration & Mean ATE & $\Delta$ vs default \\
\midrule
\rowcolor{gray!12}
Default ($\mathcal{R}{=}\{3,5\}$, $W{=}60$, $\tau{=}0.7$) & \textbf{22.61} & --- \\
\midrule
\multicolumn{3}{l}{\emph{Coarse stride set $\mathcal{R}$}} \\
$\mathcal{R}{=}\{3\}$ (drop stride 5) & 24.52 & $+1.91$ \\
$\mathcal{R}{=}\{5\}$ (drop stride 3) & 27.34 & $+4.73$ \\
$\mathcal{R}{=}\{3,5,7\}$ (add stride 7) & 22.90 & $+0.29$ \\
\midrule
\multicolumn{3}{l}{\emph{Loop-density window $W$}} \\
$W{=}30$ & 22.65 & $+0.04$ \\
$W{=}90$ & 22.65 & $+0.04$ \\
\midrule
\multicolumn{3}{l}{\emph{Adaptive direction threshold $\tau$}} \\
$\tau{=}0.3$ & 23.09 & $+0.48$ \\
$\tau{=}0.5$ & 22.85 & $+0.24$ \\
\bottomrule
\end{tabular}
\end{table}

\paragraph{Coarse stride set.}
Dropping either coarse stride degrades the mean, and stride 3 is the more important of the two.
Removing stride 3 is the most damaging change at $+4.73$\,m, while removing stride 5 still costs $+1.91$\,m, so both strides contribute and we keep them.
Adding stride 7 on top of the default slightly hurts the mean ($+0.29$\,m).
The wider frame gaps at stride 7 may be out of distribution for VGGT or may over-constrain the pose graph, so we keep $\mathcal{R}{=}\{3,5\}$.

\paragraph{Loop-density window.}
Varying $W$ from the default 60 to 30 or 90 frames changes the mean by only $0.04$\,m, so the adaptive direction policy is robust to the window size, and we keep $W{=}60$ as the default.

\paragraph{Adaptive direction threshold.}
Lowering $\tau$ from the default $0.7$ monotonically raises the mean, to $22.85$\,m at $\tau{=}0.5$ and $23.09$\,m at $\tau{=}0.3$, so $\tau{=}0.7$ is the best setting and we adopt it as the default.

\section{Limitations}
Our framework inherits limitations from both the VGGT backbone and the loop-closure detector. Because it operates at the graph level, it does not directly denoise the VGGT predictions themselves. Severe backbone errors can therefore still be propagated through the cross-scale skip edges. The adaptive anchoring policy is also limited by the reliability of loop detection, and missed loop closures in visually ambiguous or textureless scenes can lead to suboptimal anchor selection.
\methodname requires additional VGGT runs on coarse-stride chunks, increasing inference time.
Like other VGGT-based reconstruction methods, it assumes a predominantly static scene, and handling dynamic objects is orthogonal to our contribution.

\section{Additional qualitative results}
\label{sec:appendix_qualitative}
\begin{figure}[!htbp]
\centering
\includegraphics[width=\linewidth]{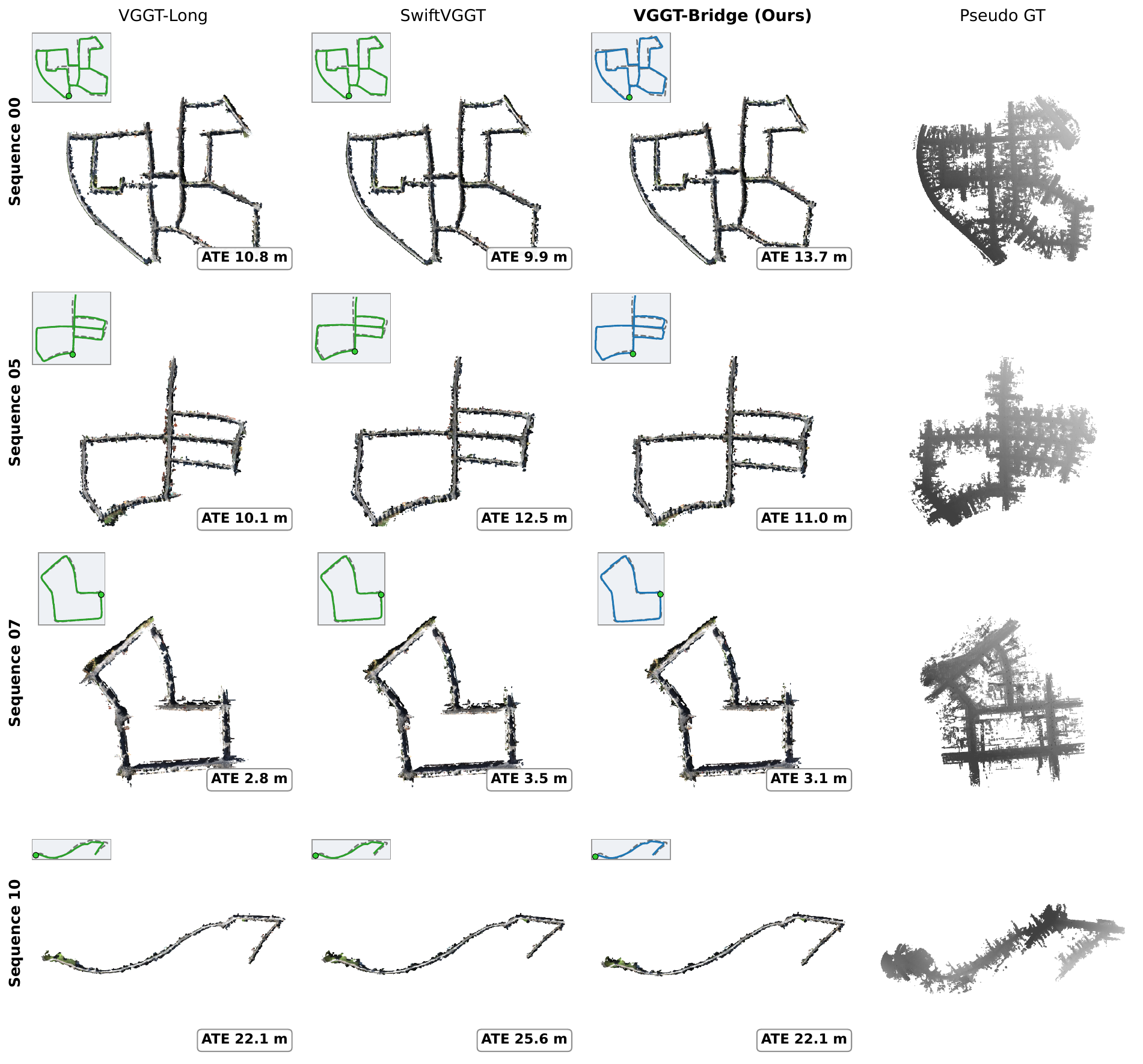}
\caption{\textbf{Additional camera trajectory and dense 3D reconstruction comparison on KITTI Odometry.}}
\label{fig:appendix_pcd}
\end{figure}

\begin{figure}[!htbp]
\centering
\includegraphics[width=\linewidth]{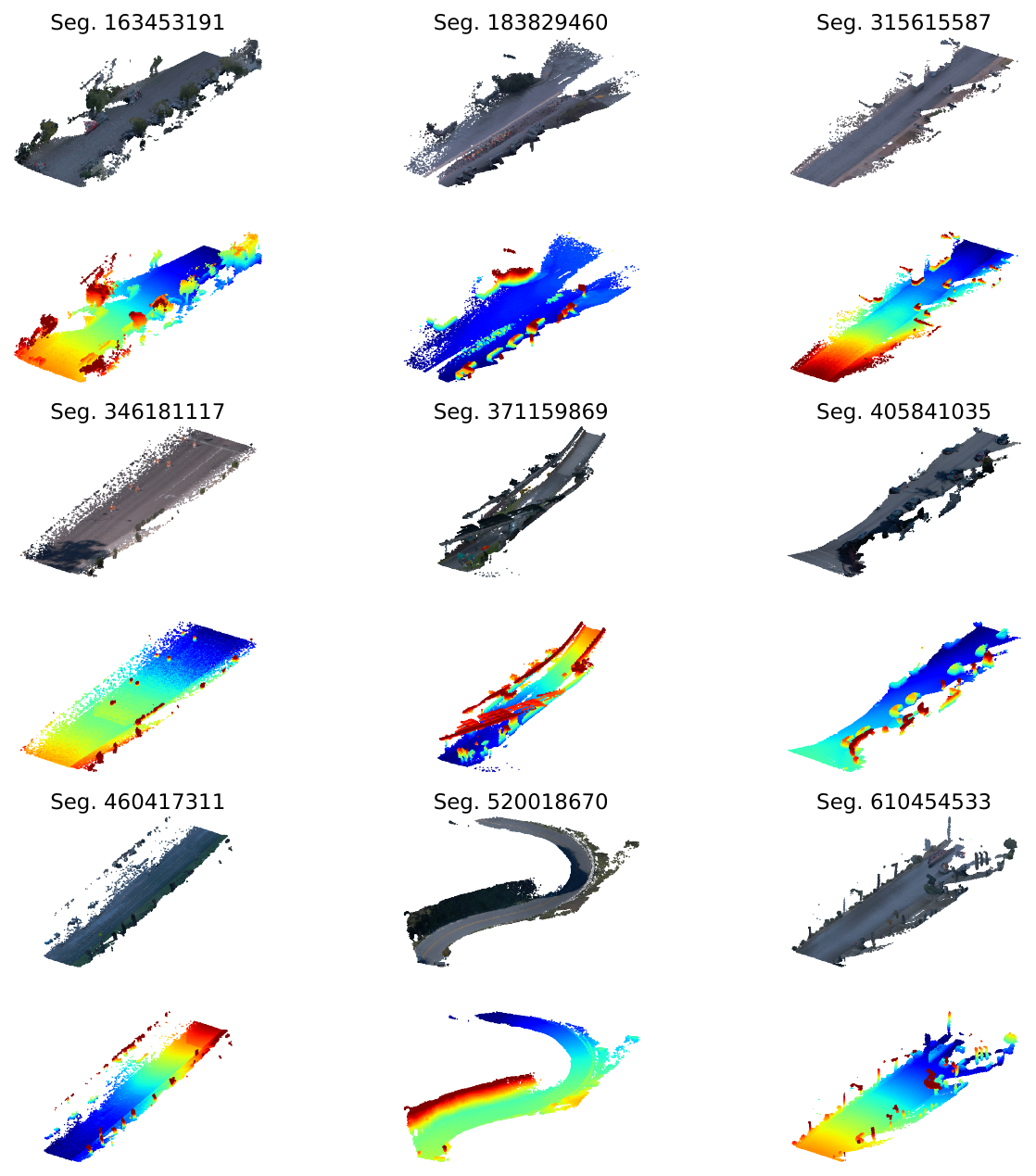}
\caption{\textbf{Visualization of 3D reconstruction on Waymo Open.}
\methodname{} reconstructions shown in RGB color (top) and colored by height, with red for low and blue for high (bottom).}
\label{fig:appendix_waymo}
\end{figure}

\begin{figure}[!htbp]
\centering
\includegraphics[width=\linewidth]{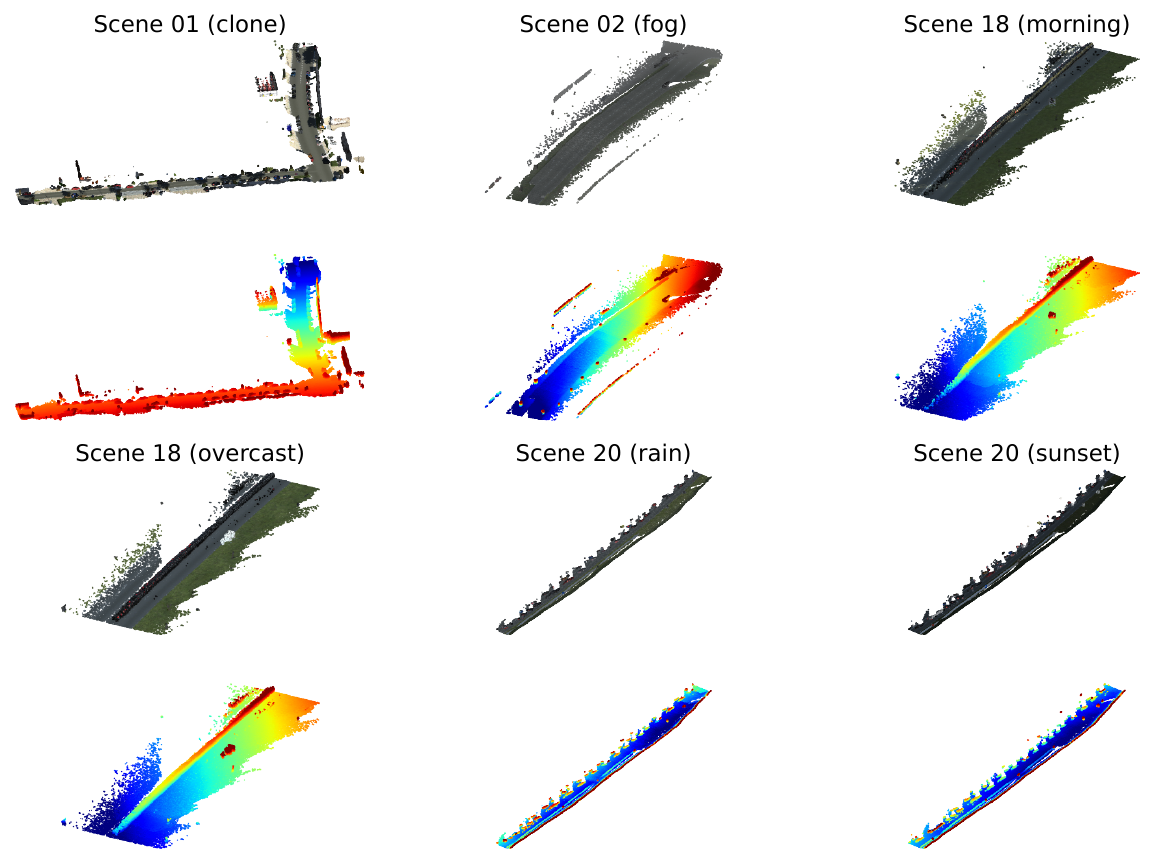}
\caption{\textbf{Visualization of 3D reconstruction on Virtual KITTI.}
\methodname{} reconstructions shown in RGB color (top) and colored by height, with red for low and blue for high (bottom).}
\label{fig:appendix_vkitti}
\end{figure}

We provide qualitative dense reconstruction results on the KITTI Odometry~\cite{kitti_odometry}, Waymo Open~\cite{waymo}, and Virtual KITTI~\cite{vkitti} datasets in \Cref{fig:appendix_pcd,fig:appendix_waymo,fig:appendix_vkitti}.
For KITTI Odometry we compare the aggregated point clouds of \methodname{} against VGGT-Long and SwiftVGGT in RGB color, while for Waymo Open and Virtual KITTI we show the \methodname{} reconstructions in both RGB color and a height-based color map.
These results show that \methodname{} produces high-quality dense 3D reconstructions across diverse large-scale driving scenarios, consistent with its tracking accuracy.

\end{document}